\documentclass{article}
\usepackage{float,flafter,placeins,etoolbox}
\floatstyle{ruled}
\newfloat{algorithm}{tbp}{loa}
\floatname{algorithm}{Algorithm}
\floatstyle{plain}
\usepackage{iclr2027_conference,times}
\usepackage{amsmath,amssymb,booktabs,graphicx,microtype}
\usepackage[hidelinks]{hyperref}
\usepackage{xurl}
\usepackage[table]{xcolor}
\usepackage{tabularx,multirow,longtable}
\definecolor{TableHeader}{HTML}{F0F2F5}
\definecolor{TableStudent}{HTML}{EAF2F8}
\definecolor{TableTeacher}{HTML}{F4F8FB}
\definecolor{TableBest}{HTML}{D5E5F1}
\definecolor{TableStudentBand}{HTML}{EDF3FC}
\definecolor{TableTeacherBand}{HTML}{FFF3E3}
\definecolor{TableRoleStudent}{HTML}{2453CC}
\definecolor{TableRoleTeacher}{HTML}{995500}
\definecolor{TableGain}{HTML}{087B55}
\definecolor{TableLoss}{HTML}{B63E52}
\definecolor{TableInterval}{HTML}{657080}
\definecolor{TableGainBand}{HTML}{EEF7F2}
\definecolor{TableOursStudent}{HTML}{DFEBFA}
\definecolor{TableOursTeacher}{HTML}{FFEDD4}
\newcommand{\compacttablecite}[1]{{\fontsize{6.2}{7}\selectfont\citep{#1}}}
\newcommand{\tablefocus}[2]{{\setlength{\fboxsep}{.5pt}\setlength{\fboxrule}{.35pt}\fcolorbox{#1}{white}{#2}}}
\newcommand{\tablehead}[1]{\textbf{#1}}
\newcommand{\tablecite}[1]{{\scriptsize\citep{#1}}}
\newcommand{\bestcell}[1]{\cellcolor{TableBest}\textbf{#1}}
\newcommand{\resulttablesetup}{%
  \centering\footnotesize
  \setlength{\tabcolsep}{3pt}%
  \renewcommand{\arraystretch}{1.16}%
}
\usepackage{needspace}
\AtBeginEnvironment{figure}{\setlength{\parskip}{0pt}}
\AtBeginEnvironment{table}{\setlength{\parskip}{0pt}}
\makeatletter
\renewcommand{\section}{\@startsection{section}{1}{\z@}
  {-1ex plus -.2ex minus -.1ex}{.4ex plus .1ex}
  {\large\sc\raggedright}}
\renewcommand{\subsection}{\@startsection{subsection}{2}{\z@}
  {-.8ex plus -.2ex minus -.1ex}{.3ex plus .1ex}
  {\normalsize\sc\raggedright}}
\renewcommand{\subsubsection}{\@startsection{subsubsection}{3}{\z@}
  {-.7ex plus -.2ex minus -.1ex}{.2ex plus .1ex}
  {\normalsize\sc\raggedright}}
\renewcommand{\paragraph}{\@startsection{paragraph}{4}{\z@}
  {.5ex plus .1ex minus .1ex}{-1em}{\normalsize\bf}}
\makeatother
\newcommand{\compactbodyspacing}{%
  \setlength{\parskip}{0.5pt plus .3pt minus .3pt}%
  \setlength{\abovedisplayskip}{2pt plus .5pt minus .5pt}%
  \setlength{\belowdisplayskip}{2pt plus .5pt minus .5pt}%
  \setlength{\abovedisplayshortskip}{0pt plus 1pt}%
  \setlength{\belowdisplayshortskip}{1pt plus .5pt minus .5pt}%
}

\newcommand{\method}{\textsc{SkillCycle}}
\title{\normalfont SkillCycle: Evolving Skills through Internalization}
\author{%
  \normalfont\small
  Ling Li\textsuperscript{1}, Qiuyu Shen\textsuperscript{2},
  Zheng Jiang\textsuperscript{1}, Qinwei Ma\textsuperscript{1},
  Yuxuan Liu\textsuperscript{1}, Zhidong Deng\textsuperscript{1}\\[-1pt]
  \normalfont\small
  \textsuperscript{1}Tsinghua University\quad
  \textsuperscript{2}Northwestern Polytechnical University
}
\date{}
\iclrfinalcopy
\hypersetup{pdftitle={SkillCycle: Evolving Skills through Internalization},pdfauthor={Ling Li, Qiuyu Shen, Zheng Jiang, Qinwei Ma, Yuxuan Liu, Zhidong Deng}}

\begin{document}
\raggedbottom
\maketitle
\lhead{Preprint}
\compactbodyspacing

\begin{abstract}
Internalizing external skills changes a language agent's capabilities and,
with them, the value of its remaining guidance: rules can become redundant,
misleading, or insufficient for newly encountered decisions. This creates
a coupled problem of learning from skills and adapting the skills that
supervise further learning.
We introduce \textbf{\method{}}, a framework for co-evolving agent policies
and skill banks through a feedback loop between skill internalization and
rule revision. Our central contribution is to give distillation feedback
a second role: token-level contextual differences help locate rules for
inspection, while interaction outcomes guide edits to their content and
applicability. SkillCycle alternates between two phases: policy learning
with a fixed skill bank and router, and rule revision with a frozen policy.
Candidate edits undergo rule-level and whole-bank environment comparisons
before they guide the next learning cycle.
On WebShop, SkillCycle with a 3B model achieves a success rate of
$\mathbf{74.74\%}$ and a score of $\mathbf{88.37}$ without inference-time
skill inputs, representing relative improvements of $0.73\%$ and $3.96\%$
over the state-of-the-art (SOTA) model, respectively. In Cycle~3 ablations
on ALFWorld and WebShop, SkillCycle's no-skill success rates improve by
$10.18\%$ and $18.11\%$ relative to a static skill bank, and by $2.41\%$
and $2.50\%$ relative to a single bank update, respectively. These results
show that continually revising skill guidance as the agent's capabilities
change helps transform external skills into policy capabilities that
require no skill inputs at inference. We will release code, configurations,
skill banks, and evaluation protocols.
\end{abstract}
\begin{figure}[!t]
\centering
\includegraphics[width=\linewidth]{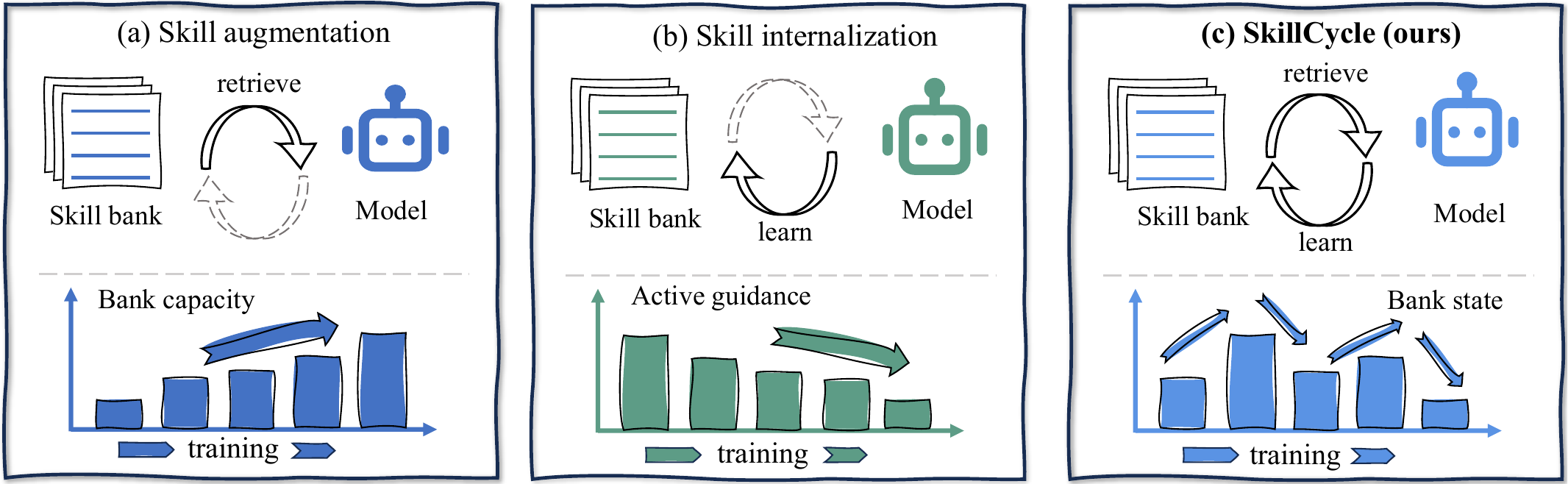}
\par\nointerlineskip
\caption{Comparison between skill augmentation and skill internalization.
Existing approaches (a) augment the policy with an external skill bank or (b) internalize skills into the policy to reduce reliance on external guidance.
(c) Our proposed SkillCycle couples skill internalization with evidence-driven skill-bank revision, co-evolving the policy and the bank for skill-free inference.
Bars and arrows illustrate conceptual trends, not measured trajectories.}
\label{fig:concept}
\end{figure}
\section{Introduction}
\label{sec:introduction}
External skills help language agents act on observations \citep{yao2022react}
through executable libraries \citep{wang2023voyager}, induced workflows
\citep{wang2024awm}, and accumulated reflections
\citep{shinn2023reflexion,zhao2023expel}. Skill augmentation supplies this
guidance in context and can evolve alongside the policy
\citep{xia2026skillrl,shi2026skill1}; skill internalization transfers its benefits into
parameters to reduce retrieval dependence. Skill0 withdraws context using
on-policy helpfulness \citep{lu2026skill0}, while Skill0.5 combines
internalization with difficulty-aware skill utilization at inference time \citep{zhu2026skill05}.

Learning also changes what guidance the policy needs. A previously useful
rule may become redundant, mislead the updated policy, or fail to cover
newly encountered states. Withdrawing guidance addresses continued reliance
on it, but does not determine how its content or applicability should evolve.
We therefore ask: \emph{can feedback from internalization guide rule revisions
that improve subsequent learning without inference-time skills?}
Figure~\ref{fig:concept} compares these approaches.

The key difficulty is distinguishing mastery from ineffective guidance.
A shrinking Teacher--Student gap can reflect successful learning or a skill
that helps neither policy. Contextual self-distillation
\citep{zhao2026opsd,yang2026opid} supplies token-level log-probability
differences that locate where guidance changes a sampled response.
These differences identify decisions to inspect; task outcomes establish
whether behavior succeeds. Together, they enable finer diagnosis than an aggregate helpfulness score; rule-removal rescoring then identifies influential instructions.

We introduce \method{}, a framework that co-evolves agent policies and
skill banks through a feedback loop between skill internalization and rule
revision. SkillCycle treats the bank as training memory that is continually
revised as the agent's capabilities change, alternating between two phases.
During internalization, the bank and critical-first router remain fixed,
while skill-conditioned supervision and environment rewards train a no-skill
Student. During revision, the Student is frozen; token-level feedback,
routing records, and interaction trajectories identify rules for inspection
and guide candidate edits to their content and applicability. Verified edits
are committed to the bank to supply updated guidance for the next learning
cycle.

The central innovation is to give distillation feedback a second role:
beyond supervising policy learning, it provides diagnostic evidence for
improving the skill guidance itself. Token-level contextual differences
locate decisions to inspect, rule-removal rescoring identifies the relevant
rules, and interaction outcomes inform whether those rules are effective
and how they should be revised. To assess candidate edits, SkillCycle holds
the policy fixed and conducts paired environment comparisons first at the
rule level and then at the whole-bank level, testing both local benefits
and effects on the assembled guidance. Only edits that pass both levels
of verification are accepted. This process turns feedback from policy
learning into improved guidance for subsequent learning, allowing the bank
to adapt to the agent's changing capabilities and needs while supporting
the transfer of external skills into policy capabilities that require no
skill inputs at inference.

Our contributions are fourfold:
\textbf{(1) Policy--skill co-evolution.}
SkillCycle alternates skill internalization and rule revision, adapting
supervision to the evolving policy for inference without skill inputs.
\textbf{(2) State-aware, critical-first routing.}
Our router matches applicable skills to observable states, prioritizes
critical rules, and records routing decisions to support subsequent revision.
\textbf{(3) Feedback-driven, verified revision.}
Token-level distillation feedback and interaction outcomes guide rule
diagnosis and editing; paired rule-level and whole-bank environment tests
verify candidate changes.
\textbf{(4) Extensive experiments and superior performance.}
Comparisons, ablations, multi-cycle evaluations, and API-model studies on
ALFWorld and WebShop validate SkillCycle, which surpasses WebShop SOTA in
success rate and score without inference-time skills and outperforms static
and single-refresh banks.

Related work is discussed in Appendix~\ref{app:related-work}.
\section{Method: SkillCycle}
\label{sec:method}
\label{sec:architecture}
\begin{figure}[t]
\centering
\includegraphics[width=\linewidth]{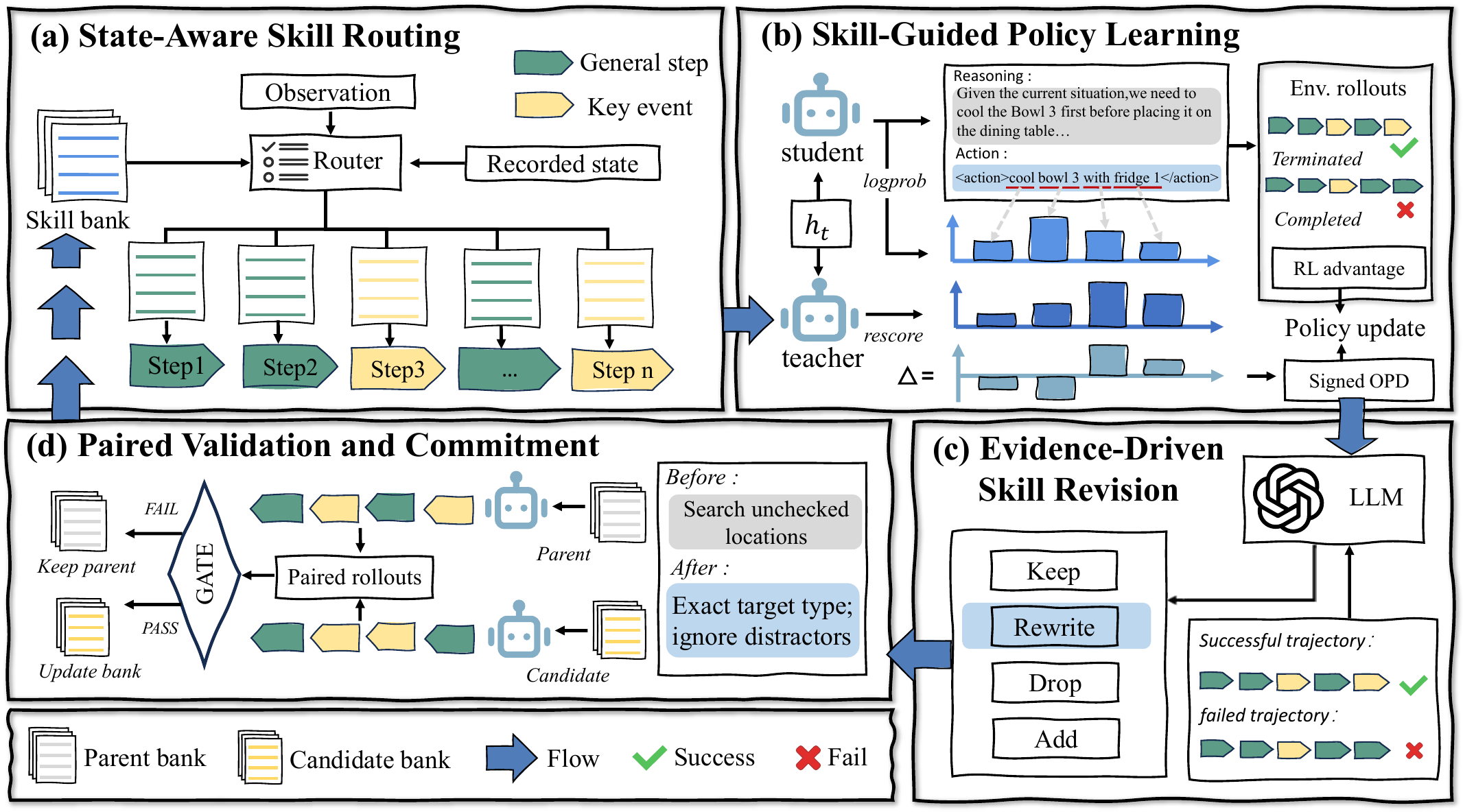}
\par\nointerlineskip
\caption{SkillCycle architecture.
(a) State-aware routing selects skill guidance from the bank.
(b) With the bank and router frozen, a shared-weight, skill-conditioned
Teacher rescores the no-skill Student's responses;
signed OPD signals and environment feedback guide policy learning.
(c) OPD diagnostics and trajectory outcomes inform offline \textsc{Keep}, \textsc{Rewrite},
\textsc{Drop}, and \textsc{Add} proposals.
(d) Paired rollouts under a frozen policy verify candidate revisions;
accepted updates guide the next learning cycle.
WebShop additionally uses verified selective Teacher demonstrations.
Here, $h_t$ denotes the task instruction and observable interaction history.
Bars and trajectories are schematic.}
\label{fig:cycle}
\end{figure}

\method{} treats the skill bank as an editable source of supervision for
a policy that acts without skill context. Figure~\ref{fig:cycle} summarizes
the cycle: state-aware routing supplies guidance for policy learning
(a,b), and trajectory evidence drives skill revision and paired validation
(c,d). The validated bank supplies revised skill context for supervision in the next learning cycle.

\hyphenpenalty=10000\exhyphenpenalty=10000
\definecolor{FigInk}{HTML}{343941}
\definecolor{FigMuted}{HTML}{756C64}
\definecolor{FigPolicy}{HTML}{B55B45}
\definecolor{FigSkill}{HTML}{D99332}
\definecolor{FigSkillText}{HTML}{A06B24}
\definecolor{FigBaseline}{HTML}{846D99}
\definecolor{FigAccept}{HTML}{547C64}
\definecolor{FigReject}{HTML}{A64C56}
\definecolor{FigLine}{HTML}{D7CEC4}
\definecolor{FigNeutralBg}{HTML}{F6F3EF}
\definecolor{FigAmberBg}{HTML}{F7EEDC}
\definecolor{FigLavenderBg}{HTML}{EEE9F5}
\definecolor{FigRoseBg}{HTML}{F6E8E2}
\definecolor{FigLavender}{HTML}{B4A5C6}
\definecolor{FigAmber}{HTML}{EAB36B}
\definecolor{FigRose}{HTML}{DCA6A0}
\colorlet{FigPolicyBg}{FigPolicy!5}
\colorlet{FigSkillBg}{FigSkill!5}
\subsection{Problem Setup}
\label{sec:problem}
Given instruction-following tasks $x\sim\mathcal D$, a language policy
generates response $y_t$ from observable history $h_t$, and the environment
executes its parsed action. A bank $B$ contains reusable skill documents
with identifiable rules and applicability metadata. The router selects
$k_t=R(h_t;B)\in B\cup\{\varnothing\}$ using only observable information;
documents supply context, while rules and metadata can be revised.
Following contextual self-distillation \citep{zhao2026opsd}, the no-skill
Student $\pi^S_\theta$ and skill-conditioned Teacher $\pi^T_\theta$ share
parameters and differ only in whether $k_t$ is included in context.
The Teacher receives no hidden states, future observations, or task answers.

Let $U_x(\tau)$ denote trajectory success or task score. Starting from
$(\theta_0,B_0,R_0)$, we seek to improve the final Student's expected utility
under a fixed training budget:
\begin{equation}
 J^S_{\mathcal D}(\theta_C)=
 \mathbb E_{x\sim\mathcal D,\,\tau\sim\pi^S_{\theta_C}(\cdot\mid x)}
 [U_x(\tau)].
 \label{eq:student-objective}
\end{equation}
The trajectory distribution here excludes skill context. Bank edits affect
this objective through subsequent policy learning.
Skill-conditioned evaluation separately measures guidance utility
at a fixed checkpoint. Training tasks support learning and
proposals; designated development tasks support acceptance and model
selection, with final-test tasks excluded from both. Appendix~\ref{app:interaction}
formalizes the interaction and performance measures.

Let $H_c=(B_c,R_c)$ denote the committed bank and routing declarations
after revision $c$. Given training and acceptance sets
$\mathcal D_{\mathrm{train}}$ and $\mathcal D_{\mathrm{acc}}$, the cycle is
\begin{equation}
\begin{aligned}
 (\theta_c,E_c)&=\operatorname{Internalize}_{\omega}
       (\theta_{c-1},H_{c-1};\mathcal D_{\mathrm{train}}),
       &&1\leq c\leq C,\\
 H_c&=\operatorname{Revise}_{\omega}
       (H_{c-1};\theta_c,E_c,\mathcal D_{\mathrm{acc}}),
       &&1\leq c<C,
\end{aligned}
\label{eq:cycle}
\end{equation}
where $\omega$ specifies update counts, budgets, and acceptance protocols,
and $E_c$ records the training evidence. Each segment learns with
$H_{c-1}$ fixed, then tests bank edits with $\theta_c$ fixed.
Each segment ends with a no-skill evaluation of Eq.~\eqref{eq:student-objective}; the last omits revision (Algorithm~\ref{alg:skillcycle}, Appendix~\ref{app:algorithm}).

This alternation separates learning from changes to its supervision.
Within a segment, the bank defines a stable source of skill context while
the same-weight Teacher tracks the Student; it has no independent optimizer.
Between segments, holding the policy fixed makes candidate comparisons
specific to the proposed guidance change. A revision can therefore be
tested before it influences further parameter updates. Acceptance measures
immediate guidance utility at $\theta_c$, while the next no-skill evaluation
measures whether training with the committed bank improves the learner.

\subsection{Skill Internalization}
\label{sec:internalization}
\label{sec:routing}
At each decision, the router (Figure~\ref{fig:cycle}(a)) prioritizes applicable critical-event skills,
falls back to general workflows when none is eligible, and abstains on
unresolved ties. It uses observable facts to select Teacher guidance;
the Student acts without skills. Routing code is fixed, while declarations
can change between cycles. Eligibility, ranking, and repeat suppression
are detailed in Appendix~\ref{app:detail-routing}.

Eligibility precedes priority: active documents must satisfy observable
preconditions and pass blocking and repeat checks; unknown facts cannot
establish applicability. Eligible documents are ranked by priority, then
predicate specificity. A critical-event tie causes abstention rather than
general fallback. Episode-local records suppress repeated bindings until
confirmed progress; the policy selects the action rather than executing a router-prescribed command.

For policy learning (Figure~\ref{fig:cycle}(b)), at update $u$ of cycle $c$,
the no-skill Student samples trajectories
$\mathcal T_{c,u}$ using $\theta_{c,u}$. The same-weight Teacher rescores
these responses with routed skills and identical response prefixes.
For token $j$ at interaction step $t$, the distillation signal is
\begin{equation}
 g_{t,j}=\operatorname{sg}\!\left[
 \log\pi^T_{\theta_{c,u}}(y_{t,j}\mid h_t,k_t,y_{t,<j})
 -\log\pi^S_{\theta_{c,u}}(y_{t,j}\mid h_t,y_{t,<j})
 \right].
 \label{eq:opd-gap}
\end{equation}
Here $\operatorname{sg}$ stops gradients. With matched weights and prefixes,
$g_{t,j}>0$ means skill context increases the sampled token's probability;
$g_{t,j}<0$ means it decreases that probability. Both signs are retained.
These preferences need not be correct; environment outcomes supply task-level feedback.

Rescoring aligns supervision with learner-visited states
\citep{ross2011dagger,agarwal2024onpolicy}; routing abstention gives zero gap while
environment feedback remains active.

Let $\ell_{t,j}(\theta)$ be the current no-skill Student log probability,
and $m^o$ select OPD-applied response tokens, excluding appended demos.
The objective separates advantage injection and auxiliary loss:
\begin{equation}
\begin{aligned}
 A^{\mathrm{mix}}_{t,j}&=\operatorname{sg}(A^{\mathrm{env}}_t)
   +\eta m^o_{t,j}\operatorname{clip}(g_{t,j},-c_g,c_g),\\
 \mathcal L&=\mathcal L_{\mathrm{PPO}}(A^{\mathrm{mix}})
 +\lambda_o\mathcal L_{\mathrm{OPD}}+\lambda_d\mathcal L_{\mathrm{demo}}
 +\beta\mathcal L_{\mathrm{ref}}-\alpha\mathcal H,\\
 \mathcal L_{\mathrm{OPD}}&=-\mathcal N_{m^o}(g\ell(\theta)),\qquad
 \mathcal L_{\mathrm{demo}}=-\mathcal N_{m^d}(\ell^z(\theta)).
\end{aligned}
 \label{eq:webshop-objective}
\end{equation}
Here $\ell^z$ scores tokens of a verified alternative response $z$ under
the Student, $m^d$ selects supervision tokens, and $\mathcal N$ is the
run-specific masked reduction. Gradients flow through
the current Student log probabilities; $g$ and the rollout policy remain
fixed during optimization. PPO \citep{schulman2017ppo} uses ratio bounds $[0.8,1.2]$ and dual
clip~3; demo rows are excluded from PPO and its reference/entropy regularizers.
For WebShop Cycle~3, $(\eta,\lambda_o,\lambda_d)$ is
$(0,0.001,0.1)$ for 1.7B and $(0,0,0.1)$ for 3B; the latter uses OPD to
select supervision without an auxiliary OPD loss. The ALFWorld 3B Cycle~3
configuration injects signed gaps with $\eta=0.1$ and no auxiliary loss.
Table~\ref{tab:audited-training-configs} lists audited configurations;
Appendix~\ref{app:training} specifies group-centered environment advantages,
token masks, reductions, synchronization, and configuration scope.
Evidence $E_c$ records sampled responses, log probabilities, routing decisions, token masks, and observed outcomes.

\subsection{Evidence-Guided Skill Revision}
\label{sec:screening}
Outcomes identify success, while local probability comparisons help locate
influential instructions (Figure~\ref{fig:cycle}(c)). With $\theta_c$
fixed, diagnosis inspects action tokens: absolute OPD gaps prioritize
decisions; invalid and successful terminal actions supply directional
labels. Intermediate actions remain unlabeled. Rule influence compares
the same response with and without exposed rule $r$:
\begin{equation}
 L_c(t;k)=\frac{1}{n_t}\sum_jm^a_{t,j}
 \log\pi^T_{\theta_c}(y_{t,j}\mid h_t,k,y_{t,<j}),\qquad
 d_{t,r}=L_c(t;k_t)-L_c(t;k_t\setminus r).
 \label{eq:rule-influence}
\end{equation}
Here $m^a=m^{\mathrm{resp}}\odot m^{\mathrm{applied}}\odot m^{\mathrm{action}}$ selects
action tokens with valid OPD scores, and $n_t=\sum_jm^a_{t,j}>0$.
The inspection score is $I_t=n_t^{-1}\sum_jm^a_{t,j}|g_{t,j}|$;
averaging over selected tokens prevents response length from inflating it.
Both rule contexts use the same frozen scorer, so $d_{t,r}$ measures the
rule's local influence on the sampled action.
The full and rule-removed contexts are separately rendered and rescored
with the post-segment checkpoint $\theta_c$. A score cached earlier in
training cannot substitute for either term when its parameter snapshot
differs. Fixing the sampled response isolates the likelihood change due to rule context.

Signed gaps guide learning on valid response tokens; action-only magnitudes
prioritize diagnosis without implying success. Diagnosis retains labeled
decisions and selects high-disagreement unlabeled records with episode and
task caps to limit repeated evidence. The proposer uses training-side
records, without access to acceptance-task contents or final-test trajectories.

An offline proposer uses this evidence to suggest \textsc{Keep},
\textsc{Rewrite}, \textsc{Drop}, or \textsc{Add} operations. Failed decisions
without a matching rule can motivate additions, allowing the bank to expand
its coverage as well as correct existing guidance. For a rewrite
$q$, let $\delta_t(q)=L_c(t;k_t^{(q)})-L_c(t;k_t)$ be its change in
action log likelihood, where $k_t^{(q)}$ replaces the affected rule.
Define $S(q)=\operatorname{HMean}[\ell_t\delta_t(q)]$ on labeled records
and $A(q)=\operatorname{HMean}[|\delta_t(q)|]$ on all selected records.
The labels are $-1$ for invalid actions and $+1$ for successful terminal
actions; other actions contribute magnitude information only.
$\operatorname{HMean}$ averages within tasks, then across tasks and seeds.
Rewrites pass screening when $A(q)>\varepsilon$ and, where labels exist,
$S(q)\geq0$: they must produce a nontrivial likelihood change with a
nonnegative mean outcome-aligned effect on labeled actions. This average
criterion permits individual regressions, which motivate subsequent
environment verification. The scorer's default is $\varepsilon=10^{-4}$;
ranking and configurable evidence caps appear in Appendix~\ref{app:screening}.
Related text-revision methods include Self-Refine \citep{madaan2023selfrefine}
and TextGrad \citep{yuksekgonul2024textgrad}. Here, revisions change a
persistent source of training supervision and require environment verification.
Appendices~\ref{app:detail-screening} and~\ref{app:screening} detail
evidence selection and rescoring.

\subsection{Verified Bank Updates}
\label{sec:bank-update}
Paired validation (Figure~\ref{fig:cycle}(d)) evaluates proposed changes before
committing them to the bank. In sequential verification, $H^{(0)}=H_{c-1}$
and each valid edit $q_i$ produces a trial
$\widetilde H^{(i)}=T_{q_i}(H^{(i-1)})$. With the policy fixed, matched
acceptance tasks and decoding seeds estimate its effect:
\begin{equation}
 \widehat\Delta(q_i)=
 \sum_{(x,\xi)\in\mathcal P_{q_i}}
 w_{x,\xi}\left[U_x(\tau^{\widetilde H^{(i)}}_{x,\xi})
       -U_x(\tau^{H^{(i-1)}}_{x,\xi})\right].
\label{eq:candidate-effect}
\end{equation}
The nonnegative weights sum to one and implement the declared metric:
equal episode weights give pooled performance, while equal family weights
give ALFWorld macro SR (Appendix~\ref{app:verification}).
The targeted gate tests edits on relevant tasks. A whole-bank gate then
compares the assembled revision with $H_{c-1}$, checking for routing
interactions or conflicting instructions. Passing commits $H_c$; failure
restores the previous bank and routing declarations.
The WebShop 3B Cycle~2--3 bank gate requires strictly more successes,
nondecreasing mean Score, no success-count regression in any monitored
attribute group, and zero API errors or interface fallbacks on matched panels.
ALFWorld's task-specific edit tests require $\sum_f(G_f-L_f)>0$ and
$G_f-L_f\geq0$ in every affected family, where $G_f$ and $L_f$ count
paired gains and losses. These are empirical acceptance criteria, not
significance tests. Run-specific panels, thresholds, and candidate-removal
comparisons appear in Appendix~\ref{app:verification}.

Targeted acceptance is provisional: individually useful edits can interact
through shared rules or routing declarations. The whole-bank test therefore
compares the assembled revision with the previously committed state,
restoring both text and routing metadata on rejection. The bank has no
prescribed growth or shrinkage schedule; additions expand coverage, while
rewrites and removals repair guidance for the current learner. Appendix~\ref{app:historical-cases} illustrates a rule revision.

\begin{table}[!t]
\centering\normalfont\fontsize{8}{9.2}\selectfont
\setlength{\tabcolsep}{2pt}
\renewcommand{\arraystretch}{1.04}
\setlength{\heavyrulewidth}{.65pt}
\setlength{\lightrulewidth}{.4pt}
\setlength{\cmidrulewidth}{.4pt}
\definecolor{TableHeader}{HTML}{F0F2F5}
\definecolor{TableStudent}{HTML}{EAF0F8}
\colorlet{TableOursStudent}{TableStudent}
\colorlet{TableOursTeacher}{TableStudent}
\definecolor{TableGainBand}{HTML}{F4F5F6}
\colorlet{TableRoleStudent}{black}
\colorlet{TableRoleTeacher}{black}
\colorlet{TableGain}{black}
\colorlet{TableLoss}{black}
\colorlet{TableInterval}{black}
\renewcommand{\bestcell}[1]{\textbf{#1}}
\renewcommand{\tablefocus}[2]{{\setlength{\fboxsep}{.5pt}\setlength{\fboxrule}{.35pt}\fcolorbox{black}{TableStudent}{#2}}}
\renewcommand{\tablecite}[1]{{\fontsize{6.2}{7.2}\selectfont\citep{#1}}}
\renewcommand{\compacttablecite}[1]{\tablecite{#1}}
\caption{Comparison with RL and skill-distillation methods.
C: training cycle; S: no-skill Student; T: skill-conditioned Teacher.
Baseline scores follow \citet{yang2026opid}; evaluation panels are specified
in Section~\ref{sec:benchmarks}. Bold marks maxima within each reference
block or local role; boxes mark final results. $\Delta$: last-minus-first
change (points). All metrics are on a 0--100 scale; -- denotes missing results.}
\label{tab:main-results}
\newcommand{\benchmarkhead}[1]{\textbf{#1}}
\begin{tabularx}{\linewidth}{>{\raggedright\arraybackslash}p{3.65cm}cc*{9}{>{\raggedleft\arraybackslash}X}}
\toprule
& & & \multicolumn{7}{c}{\benchmarkhead{ALFWorld}} & \multicolumn{2}{c}{\benchmarkhead{WebShop}}\\
\cmidrule(lr){4-10}\cmidrule(lr){11-12}
\rowcolor{TableHeader}
\benchmarkhead{Method} & \benchmarkhead{C} & \benchmarkhead{S/T} & \multicolumn{1}{c}{\benchmarkhead{Pick}} & \multicolumn{1}{c}{\benchmarkhead{Look}} & \multicolumn{1}{c}{\benchmarkhead{Clean}} & \multicolumn{1}{c}{\benchmarkhead{Heat}} & \multicolumn{1}{c}{\benchmarkhead{Cool}} & \multicolumn{1}{c}{\benchmarkhead{Pick2}} & \multicolumn{1}{c}{\benchmarkhead{Avg.}} & \multicolumn{1}{c}{\benchmarkhead{Score}} & \multicolumn{1}{c}{\benchmarkhead{SR}}\\
\midrule
\multicolumn{12}{l}{\textbf{(a) Qwen3-1.7B-Instruct}}\\[2pt]
Vanilla \compacttablecite{yang2025qwen3} & -- &  & 25.00 & 22.20 & 3.10 & 0.00 & 21.40 & 4.20 & 12.50 & 46.50 & 4.70\\
Skill-Prompt$^{*}$ \compacttablecite{yang2026opid} & -- &  & 10.30 & 50.00 & 16.10 & 0.00 & 0.00 & 5.00 & 9.40 & 23.00 & 2.30\\
OPSD \compacttablecite{zhao2026opsd} & -- &  & 26.30 & 33.30 & 9.10 & 0.00 & 4.50 & 5.30 & 14.10 & 47.40 & 9.30\\
GRPO \compacttablecite{shao2024grpo} & -- &  & 71.10 & 41.70 & 36.40 & 40.00 & 31.80 & 31.60 & 46.10 & 67.30 & 38.30\\
Skill-GRPO \compacttablecite{yang2026opid} & -- &  & 27.60 & 54.50 & 22.70 & 27.30 & 0.00 & 19.20 & 21.10 & 73.40 & 46.10\\
Skill-GRPO$^{*}$ \compacttablecite{yang2026opid} & -- &  & 31.40 & 42.90 & 51.90 & 8.30 & 11.50 & 7.10 & 28.10 & 80.40 & 50.00\\
GRPO+OPSD \compacttablecite{yang2026opid} & -- &  & 38.20 & 50.00 & 30.80 & 28.60 & 30.00 & 21.10 & 32.00 & 70.70 & 38.30\\
Skill-SD \compacttablecite{wang2026skillsd} & -- &  & 52.90 & 37.50 & 69.20 & \textbf{42.90} & 60.00 & 36.80 & 52.30 & \textbf{81.80} & 53.90\\
RLSD \compacttablecite{yang2026rlsd} & -- &  & 50.00 & 37.50 & 61.50 & 19.00 & 50.00 & 21.10 & 42.20 & 74.00 & 50.80\\
SDAR \compacttablecite{lu2026sdar} & -- &  & \textbf{73.50} & 25.00 & \textbf{76.90} & 33.30 & 40.00 & 36.80 & 53.90 & 76.80 & 58.60\\
OPID \compacttablecite{yang2026opid} & -- &  & 65.90 & \textbf{72.70} & 66.70 & 40.00 & \textbf{63.20} & \textbf{45.00} & \textbf{58.90} & 79.60 & \textbf{64.80}\\
\midrule
\rowcolor{TableOursStudent}
 &  & \textcolor{TableRoleStudent}{\textbf{S}} & 60.61 & 49.21 & 53.33 & 61.40 & 28.07 & 22.73 & 45.89 & 82.72 & 54.69\\
\rowcolor{TableOursTeacher}
 & \multirow{-2}{*}{0} & \textcolor{TableRoleTeacher}{\textbf{T}} & 68.18 & \textcolor{TableRoleTeacher}{\textbf{55.56}} & 53.33 & 54.39 & 45.61 & 19.70 & 49.46 & 83.74 & 59.38\\
\rowcolor{TableOursStudent}
 &  & \textcolor{TableRoleStudent}{\textbf{S}} & \textcolor{TableRoleStudent}{\textbf{73.15}} & \textcolor{TableRoleStudent}{\textbf{50.00}} & \textcolor{TableRoleStudent}{\textbf{55.07}} & \textcolor{TableRoleStudent}{\textbf{72.22}} & \textcolor{TableRoleStudent}{\textbf{37.68}} & \textcolor{TableRoleStudent}{\textbf{35.90}} & \tablefocus{TableRoleStudent}{\textcolor{TableRoleStudent}{\textbf{54.00}}} & \tablefocus{TableRoleStudent}{\textcolor{TableRoleStudent}{\textbf{84.65}}} & \tablefocus{TableRoleStudent}{\textcolor{TableRoleStudent}{\textbf{59.64}}}\\
\rowcolor{TableOursTeacher}
\multirow{-4}{=}{\raggedright\textbf{\method{}}} & \multirow{-2}{*}{3} & \textcolor{TableRoleTeacher}{\textbf{T}} & \textcolor{TableRoleTeacher}{\textbf{74.07}} & 54.17 & \textcolor{TableRoleTeacher}{\textbf{65.22}} & \textcolor{TableRoleTeacher}{\textbf{72.22}} & \textcolor{TableRoleTeacher}{\textbf{50.72}} & \textcolor{TableRoleTeacher}{\textbf{32.05}} & \tablefocus{TableRoleTeacher}{\textcolor{TableRoleTeacher}{\textbf{58.07}}} & \tablefocus{TableRoleTeacher}{\textcolor{TableRoleTeacher}{\textbf{86.29}}} & \tablefocus{TableRoleTeacher}{\textcolor{TableRoleTeacher}{\textbf{70.31}}}\\
\cmidrule(lr){2-12}
\rowcolor{TableGainBand}
 &  & \textcolor{TableRoleStudent}{\textbf{S}} & \textcolor{TableGain}{\textbf{+12.54}} & \textcolor{TableGain}{\textbf{+0.79}} & \textcolor{TableGain}{\textbf{+1.74}} & \textcolor{TableGain}{\textbf{+10.82}} & \textcolor{TableGain}{\textbf{+9.61}} & \textcolor{TableGain}{\textbf{+13.17}} & \textcolor{TableGain}{\textbf{+8.11}} & \textcolor{TableGain}{\textbf{+1.93}} & \textcolor{TableGain}{\textbf{+4.95}}\\
\rowcolor{TableGainBand}
\multirow{-2}{=}{\raggedright $\Delta$ last $-$ first (points)} &  & \textcolor{TableRoleTeacher}{\textbf{T}} & \textcolor{TableGain}{\textbf{+5.89}} & \textcolor{TableLoss}{\textbf{-1.39}} & \textcolor{TableGain}{\textbf{+11.89}} & \textcolor{TableGain}{\textbf{+17.83}} & \textcolor{TableGain}{\textbf{+5.11}} & \textcolor{TableGain}{\textbf{+12.35}} & \textcolor{TableGain}{\textbf{+8.61}} & \textcolor{TableGain}{\textbf{+2.55}} & \textcolor{TableGain}{\textbf{+10.94}}\\
\midrule
\multicolumn{12}{l}{\textbf{(b) Qwen2.5-3B-Instruct}}\\[2pt]
Vanilla \compacttablecite{yang2024qwen25} & -- &  & 44.40 & 11.10 & 6.20 & 15.40 & 28.60 & 12.50 & 21.90 & 6.70 & 0.80\\
Skill-Prompt$^{*}$ \compacttablecite{yang2026opid} & -- &  & 51.70 & 66.70 & 48.40 & 0.00 & 4.30 & 10.00 & 28.90 & 0.20 & 0.80\\
OPSD \compacttablecite{zhao2026opsd} & -- &  & 48.80 & 41.70 & 16.70 & 0.00 & 15.80 & 16.70 & 28.10 & 11.30 & 3.10\\
GRPO \compacttablecite{shao2024grpo} & -- &  & 91.20 & 62.50 & 96.20 & 61.90 & 65.00 & 47.40 & 75.00 & 79.80 & 63.30\\
Skill-GRPO \compacttablecite{yang2026opid} & -- &  & 88.90 & 71.40 & 58.80 & 70.60 & 40.70 & 29.20 & 60.20 & 77.30 & 60.90\\
Skill-GRPO$^{*}$ \compacttablecite{yang2026opid} & -- &  & 94.30 & 57.10 & \textbf{100.00} & 66.70 & 73.10 & 57.10 & 80.50 & 76.30 & 66.40\\
GRPO+OPSD \compacttablecite{yang2026opid} & -- &  & \textbf{100.00} & 82.40 & 85.70 & \textbf{75.00} & 70.00 & 60.00 & 81.20 & 77.80 & 66.40\\
Skill-SD \compacttablecite{wang2026skillsd} & -- &  & 88.20 & 50.00 & 96.20 & 52.40 & 65.00 & 57.90 & 73.40 & 75.90 & 64.00\\
RLSD \compacttablecite{yang2026rlsd} & -- &  & 87.90 & 75.00 & 90.90 & \textbf{75.00} & 73.10 & 68.40 & 79.70 & 84.40 & 66.40\\
SDAR \compacttablecite{lu2026sdar} & -- &  & 97.10 & 62.50 & \textbf{100.00} & 61.90 & 75.00 & \textbf{84.20} & \textbf{84.40} & \textbf{85.00} & 68.00\\
OPID \compacttablecite{yang2026opid} & -- &  & 92.70 & \textbf{100.00} & 88.90 & 70.00 & \textbf{84.20} & 70.00 & 84.30 & \textbf{85.00} & \textbf{74.20}\\
\midrule
\rowcolor{TableOursStudent}
 &  & \textcolor{TableRoleStudent}{\textbf{S}} & 89.12 & 72.35 & 68.47 & 64.28 & 52.16 & 58.08 & 67.41 & 85.50 & 69.53\\
\rowcolor{TableOursTeacher}
 & \multirow{-2}{*}{1} & \textcolor{TableRoleTeacher}{\textbf{T}} & 86.24 & 62.37 & 77.15 & 65.42 & 65.83 & 60.47 & 69.58 & 87.02 & 73.96\\
\rowcolor{TableOursStudent}
 &  & \textcolor{TableRoleStudent}{\textbf{S}} & 85.23 & 74.36 & 82.57 & 68.41 & 63.49 & 76.54 & 75.10 & \tablefocus{TableRoleStudent}{\textcolor{TableRoleStudent}{\textbf{88.37}}} & \tablefocus{TableRoleStudent}{\textcolor{TableRoleStudent}{\textbf{74.74}}}\\
\rowcolor{TableOursTeacher}
 & \multirow{-2}{*}{3} & \textcolor{TableRoleTeacher}{\textbf{T}} & 89.53 & 75.24 & 90.81 & 75.43 & 68.29 & 88.68 & 81.33 & \tablefocus{TableRoleTeacher}{\textcolor{TableRoleTeacher}{\textbf{89.85}}} & \tablefocus{TableRoleTeacher}{\textcolor{TableRoleTeacher}{\textbf{76.82}}}\\
\rowcolor{TableOursStudent}
 &  & \textcolor{TableRoleStudent}{\textbf{S}} & \textcolor{TableRoleStudent}{\textbf{92.31}} & \textcolor{TableRoleStudent}{\textbf{79.42}} & \textcolor{TableRoleStudent}{\textbf{89.63}} & \textcolor{TableRoleStudent}{\textbf{74.27}} & \textcolor{TableRoleStudent}{\textbf{70.19}} & \textcolor{TableRoleStudent}{\textbf{89.36}} & \tablefocus{TableRoleStudent}{\textcolor{TableRoleStudent}{\textbf{82.53}}} & -- & --\\
\rowcolor{TableOursTeacher}
\multirow{-6}{=}{\raggedright\textbf{\method{}}} & \multirow{-2}{*}{4} & \textcolor{TableRoleTeacher}{\textbf{T}} & \textcolor{TableRoleTeacher}{\textbf{93.61}} & \textcolor{TableRoleTeacher}{\textbf{76.53}} & \textcolor{TableRoleTeacher}{\textbf{94.82}} & \textcolor{TableRoleTeacher}{\textbf{77.21}} & \textcolor{TableRoleTeacher}{\textbf{69.43}} & \textcolor{TableRoleTeacher}{\textbf{90.72}} & \tablefocus{TableRoleTeacher}{\textcolor{TableRoleTeacher}{\textbf{83.72}}} & -- & --\\
\cmidrule(lr){2-12}
\rowcolor{TableGainBand}
 &  & \textcolor{TableRoleStudent}{\textbf{S}} & \textcolor{TableGain}{\textbf{+3.19}} & \textcolor{TableGain}{\textbf{+7.07}} & \textcolor{TableGain}{\textbf{+21.16}} & \textcolor{TableGain}{\textbf{+9.99}} & \textcolor{TableGain}{\textbf{+18.03}} & \textcolor{TableGain}{\textbf{+31.28}} & \textcolor{TableGain}{\textbf{+15.12}} & \textcolor{TableGain}{\textbf{+2.87}} & \textcolor{TableGain}{\textbf{+5.21}}\\
\rowcolor{TableGainBand}
\multirow{-2}{=}{\raggedright $\Delta$ last $-$ first (points)} &  & \textcolor{TableRoleTeacher}{\textbf{T}} & \textcolor{TableGain}{\textbf{+7.37}} & \textcolor{TableGain}{\textbf{+14.16}} & \textcolor{TableGain}{\textbf{+17.67}} & \textcolor{TableGain}{\textbf{+11.79}} & \textcolor{TableGain}{\textbf{+3.60}} & \textcolor{TableGain}{\textbf{+30.25}} & \textcolor{TableGain}{\textbf{+14.14}} & \textcolor{TableGain}{\textbf{+2.83}} & \textcolor{TableGain}{\textbf{+2.86}}\\
\bottomrule
\end{tabularx}
\end{table}
\section{Experiments}
\label{sec:experiments}
We evaluate whether \method{} improves policies without inference-time skills,
whether repeated bank revision benefits learning, and where external guidance
remains useful. Cycle~0 denotes the outcome-only base policy before
\method{} skill-guided training. The no-skill Student is the primary outcome for internalization; the
same-weight Teacher measures the additional benefit of skill context.
Bank-update controls assess revision during training; API-model comparisons assess routing with the language model parameters fixed throughout inference.

\subsection{Benchmarks and Evaluation}
\label{sec:benchmarks}
\label{sec:evaluation}
We study Qwen3-1.7B \citep{yang2025qwen3} and Qwen2.5-3B
\citep{yang2024qwen25} on ALFWorld's six household
task families \citep{shridhar2020alfworld} and the WebShop
1,000-product subset \citep{yao2022webshop}. ALFWorld reports family success
rates (SR) and their unweighted macro average; WebShop reports SR and mean
score, all on a 0--100 scale. We use the ALFWorld and WebShop benchmark
datasets also evaluated by \citet{yang2026opid}; Table~\ref{tab:main-results}
includes their published baseline scores. Main benchmark results aggregate
three evaluation seeds on fixed 128-task panels, measuring performance
across stochastic rollouts. The WebShop records use a train-side development
panel; official-test evaluation remains outstanding.
The bank-update ablations use a separate experiment series.
Appendices~\ref{app:initialization} and~\ref{app:result-protocol} give
initialization and evaluation details; the release will link configurations, task/seed manifests, and episode outcomes to policy checkpoints and skill banks.

Within each ablation, all settings other than the stated intervention are
held fixed. Bank-update comparisons share the initial Student, training
tasks, prompts, optimizer settings, policy-update and rollout budgets, and
evaluation protocol. Skill-guided arms also share the initial bank and
router. Refresh once and \method{} use equal total proposal and verification
budgets, allocated across different revision schedules. Table~\ref{tab:ablation-core} reports this comparison (protocol: Appendix~\ref{app:controlled-protocol}).

\subsection{Main Results}
\label{sec:results}
\label{sec:implementation}
Table~\ref{tab:main-results} compares \method{} with RL, self-distillation,
and skill-based methods, including OPID, using the same model families.
Table~\ref{tab:ablation-core} isolates bank revision under matched controls.
Tables~\ref{tab:main-results-full-1p7b}--\ref{tab:main-results-7b}
provide complete cycle-wise results and the 7B reference baselines.

On WebShop, the no-skill 3B Student reaches Score 88.37 and SR 74.74\%,
exceeding all corresponding published baselines listed, including OPID's
85.00/74.20\% (+3.37 points/+0.54 pp). The 1.7B Student also attains the
highest listed Score at its scale: 84.65 versus Skill-SD's 81.80.
Section~\ref{sec:benchmarks} specifies the evaluation protocol and baseline
sources. Evaluation omits skill inputs.
Across three seeds, 3B Student WebShop SR is 74.22--75.78\% (Appendix~\ref{app:result-protocol}).

Across Cycles~0--3, the 1.7B Student gains 8.11 pp in ALFWorld macro SR
and 4.95 pp in WebShop SR, reaching 54.00\%/59.64\%.
From its first evaluated cycle, the 3B Student gains 15.12 pp on
ALFWorld (Cycles~1--4) and 5.21 pp on WebShop (Cycles~1--3).

The 3B ALFWorld Student reaches 82.53\% macro SR; Clean and Pick2 gain
21.16/31.28 pp. Its Pick2 SR of 89.36\% exceeds the listed OPID/SDAR
values of 70.00\%/84.20\%. Improvements vary by family: the 1.7B Look
and Clean gains are 0.79/1.74 pp.

\begin{figure}[t]
\centering
\includegraphics[width=\linewidth,trim=0bp 0bp 0bp 0bp,clip]{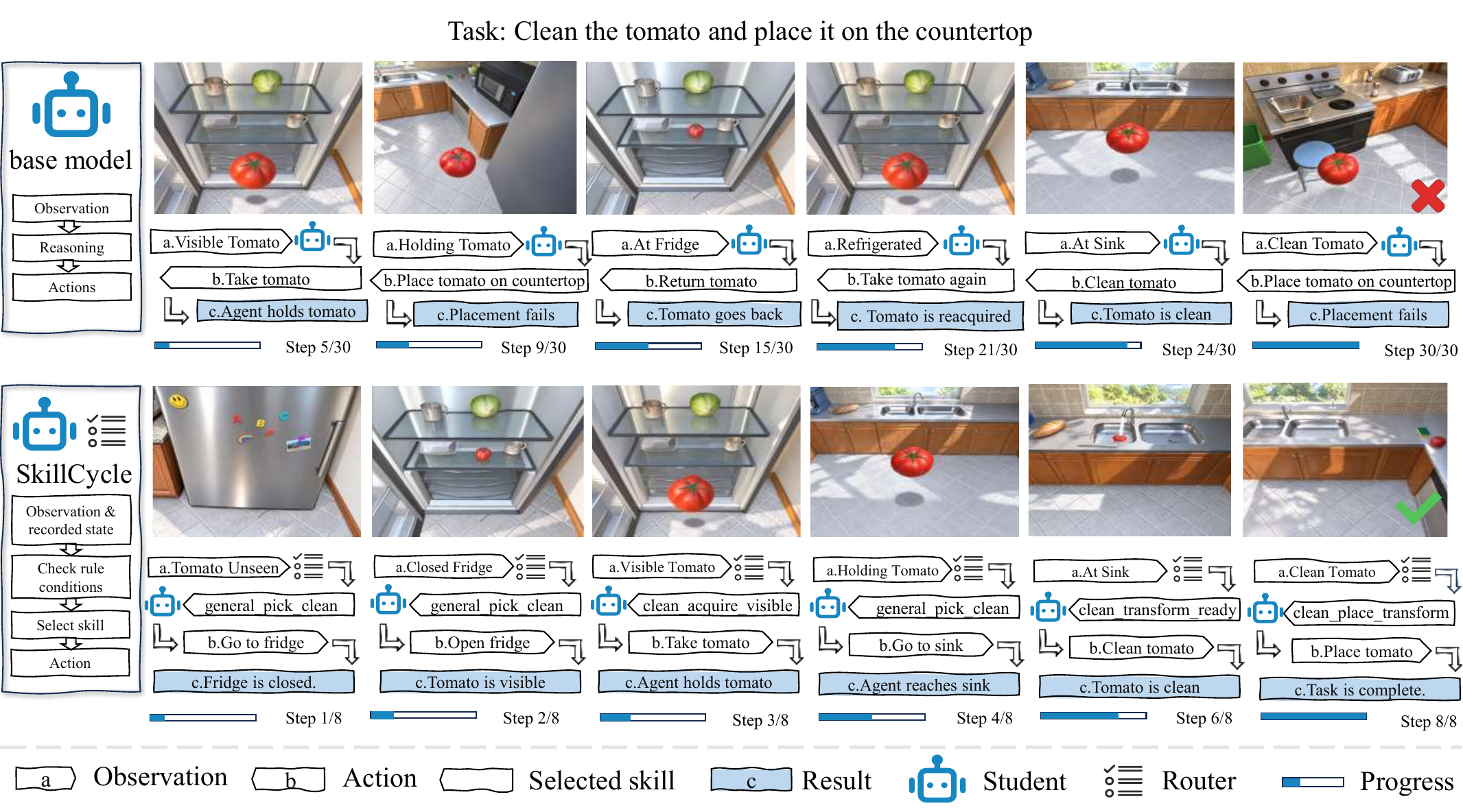}
\par\nointerlineskip
\caption{Trajectory comparison on an ALFWorld cleaning task:
the no-skill model fails at the 30-step limit (top), while the same model with
SkillCycle's routed skills succeeds in 8 steps (bottom).
Snapshots show observations, actions, outcomes, and interaction progress;
the bottom row also shows the selected skills.}
\label{fig:trajectory-comparison}
\end{figure}

The 3B ALFWorld Teacher--Student gap narrows from 6.23 pp at Cycle~3 to
1.19 pp at Cycle~4. At Cycle~3, skills still add 4.07/10.67 pp on
ALFWorld/WebShop for 1.7B and 2.08 pp on WebShop for 3B. These gaps
characterize the guidance still available beyond the learned policy.

Figure~\ref{fig:trajectory-comparison} illustrates skill-conditioned
execution on an ALFWorld cleaning task. The base model repeats retrieval
and placement attempts; \method{} changes its selected skills as the
tomato becomes visible, is acquired, and is cleaned. This case illustrates state-dependent routing; separate no-skill Student evaluations measure internalization through policy learning.
Table~\ref{tab:historical-router-cases} reports seed-wise outcomes and action
counts for two additional ALFWorld routing cases.
\subsection{Ablations and Learning Dynamics}
\label{sec:ablations}
\label{sec:baselines}
Cycle~0 anchors learning to the outcome-only initialization. At Cycle~3,
Table~\ref{tab:ablation-core} compares continued outcome-based RL (GiGPO),
Static (fixed bank), Refresh once (one bank update), and \method{}
(iterative revision). ALFWorld uses Qwen3-1.7B; WebShop uses Qwen2.5-3B.
The ALFWorld Cycle~3 Student result is 53.56\% in this controlled series
and 54.00\% in the separate benchmark series of Table~\ref{tab:main-results}.
Tables~\ref{tab:ablation-alf-student}--\ref{tab:ablation-web}
report every cycle, task family, and interval for both roles.

\begin{table}[!htbp]
\centering\normalfont\fontsize{8}{9.2}\selectfont
\setlength{\tabcolsep}{2pt}
\renewcommand{\arraystretch}{1.04}
\setlength{\heavyrulewidth}{.65pt}
\setlength{\lightrulewidth}{.4pt}
\setlength{\cmidrulewidth}{.4pt}
\definecolor{TableHeader}{HTML}{F0F2F5}
\definecolor{TableStudent}{HTML}{EAF0F8}
\colorlet{TableOursStudent}{TableStudent}
\colorlet{TableOursTeacher}{TableStudent}
\definecolor{TableGainBand}{HTML}{F4F5F6}
\colorlet{TableRoleStudent}{black}
\colorlet{TableRoleTeacher}{black}
\colorlet{TableGain}{black}
\colorlet{TableLoss}{black}
\colorlet{TableInterval}{black}
\renewcommand{\bestcell}[1]{\textbf{#1}}
\renewcommand{\tablefocus}[2]{{\setlength{\fboxsep}{.5pt}\setlength{\fboxrule}{.35pt}\fcolorbox{black}{TableStudent}{#2}}}
\renewcommand{\tablecite}[1]{{\fontsize{6.2}{7.2}\selectfont\citep{#1}}}
\renewcommand{\compacttablecite}[1]{\tablecite{#1}}
\newcommand{\corecell}[2]{\shortstack[r]{#1\\{\fontsize{6.5}{7.5}\selectfont\textcolor{TableInterval}{#2}}}}
\caption{Bank-update ablations at Cycle~3.
ALFWorld uses Qwen3-1.7B; WebShop uses Qwen2.5-3B.
Student is evaluated without skills; Teacher uses routed skills.
Cells show SR (\%) with reported intervals below; $\Delta$ denotes
percentage-point differences. Bold marks column maxima.
Interval reporting and full task/cycle results appear in
Appendix~\ref{app:ablation-records}.}
\label{tab:ablation-core}
\begin{tabularx}{\linewidth}{>{\raggedright\arraybackslash}p{3.35cm}*{4}{>{\raggedleft\arraybackslash}X}}
\toprule
& \multicolumn{2}{c}{\textbf{ALFWorld Macro SR}} & \multicolumn{2}{c}{\textbf{WebShop SR}}\\
\cmidrule(lr){2-3}\cmidrule(lr){4-5}
\rowcolor{TableHeader}
\textbf{Condition} & \textcolor{TableRoleStudent}{\textbf{Student}} & \textcolor{TableRoleTeacher}{\textbf{Teacher}} & \textcolor{TableRoleStudent}{\textbf{Student}} & \textcolor{TableRoleTeacher}{\textbf{Teacher}}\\
\midrule
GiGPO \tablecite{feng2025gigpo} & \corecell{\textbf{55.03}}{[50.03, 60.03]} & \corecell{\textbf{61.29}}{[56.29, 66.29]} & \corecell{63.54}{[58.62, 68.20]} & \corecell{64.84}{[59.94, 69.45]}\\[2pt]
Static & \corecell{48.61}{[43.61, 53.61]} & \corecell{51.82}{[46.82, 56.82]} & \corecell{63.28}{[58.35, 67.95]} & \corecell{66.41}{[61.54, 70.95]}\\[2pt]
Refresh once & \corecell{52.30}{[47.30, 57.30]} & \corecell{58.81}{[53.81, 63.81]} & \corecell{72.92}{[68.26, 77.12]} & \corecell{\textbf{76.82}}{[72.35, 80.77]}\\[2pt]
\rowcolor{TableStudent}
\textbf{\method{}} & \corecell{53.56}{[49.16, 63.81]} & \corecell{58.28}{[53.24, 63.78]} & \corecell{\textbf{74.74}}{[70.16, 78.83]} & \corecell{\textbf{76.82}}{[72.35, 80.77]}\\[2pt]
\midrule
\rowcolor{TableGainBand}
$\Delta$ vs. Static & \textcolor{TableGain}{\textbf{+4.95}} & \textcolor{TableGain}{\textbf{+6.46}} & \textcolor{TableGain}{\textbf{+11.46}} & \textcolor{TableGain}{\textbf{+10.42}}\\
\rowcolor{TableGainBand}
$\Delta$ vs. Refresh once & \textcolor{TableGain}{\textbf{+1.26}} & \textcolor{TableLoss}{\textbf{-0.53}} & \textcolor{TableGain}{\textbf{+1.82}} & \textcolor{TableInterval}{\textbf{0.00}}\\
\bottomrule
\end{tabularx}
\end{table}

\method{} reaches 53.56\%/74.74\% no-skill Student SR on ALFWorld/WebShop,
exceeding Static by 4.95/11.46 pp. A single refresh already adds 3.69/9.64 pp
over Static; continued revision contributes a further 1.26/1.82 pp.
Under matched budgets, this ordering supports both an initial revision and
redistributing revision effort across later learning stages.

The outcome-only RL comparison clarifies when guidance helps. On ALFWorld,
Static trails GiGPO by 6.42 pp; revision recovers 4.95 pp of this gap,
leaving a 1.47-pp difference. On WebShop, Static is close to GiGPO
(63.28\% versus 63.54\%), whereas \method{} reaches 74.74\%.
Revision improves four of six ALFWorld families over both bank-update
controls: Pick, Heat, Cool, and Pick2. Look and Clean remain lower.
Together with the WebShop gains, this supports a benefit across multiple
task types while identifying where skill supervision remains less effective.

\begin{figure}[!htbp]
\centering
\includegraphics[width=.93\linewidth,trim=3bp 3bp 3bp 3bp,clip]{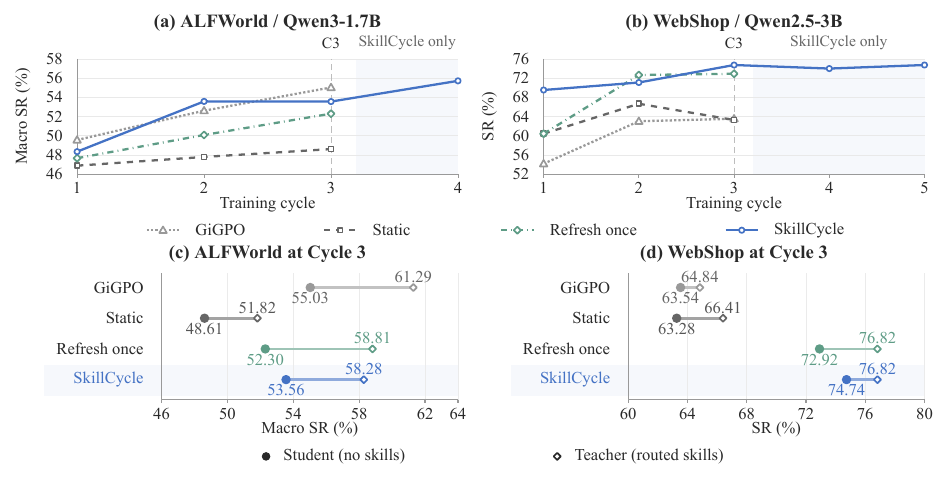}
\par\nointerlineskip
\caption{Learning dynamics and skill-conditioned performance.
(a,b) No-skill Student SR; dashed lines mark the common Cycle~3 endpoint,
shading later SkillCycle-only training. (c,d) Cycle~3 Student (circles) and
Teacher (diamonds) SR, paired by condition. Axes use separate ranges; Appendix~\ref{app:ablation-records} lists all condition-wise numerical results and their reported intervals.}
\label{fig:learning-dynamics}
\label{fig:ablation-trajectories}
\label{fig:ablation-role-comparison}
\end{figure}

Figure~\ref{fig:learning-dynamics}(c,d) contrasts learned and
skill-conditioned performance. On WebShop, \method{} and Refresh once
have identical Teacher SR (76.82\%), yet \method{} achieves higher
Student SR (74.74\% versus 72.92\%; panel~d), distinguishing immediate
guidance utility from learned capability.
The Teacher--Student gap consequently contracts from 3.90 pp with one
refresh to 2.08 pp with iterative revision. Because Teacher performance
is unchanged in this comparison, the smaller gap accompanies improved
no-skill behavior rather than a weaker Teacher. Together with the matched
revision budgets, this supports revisiting guidance as the Student changes.

With a frozen policy, bank revision improves ALFWorld macro SR from
49.43\% to 52.45\%. The WebShop series reaches 68.23\% and 67.97\%
with successive bank revisions (Table~\ref{tab:frozen-policy}). These
indices count bank updates at unchanged weights; Appendix~\ref{app:ablation-records}
gives initial configurations.
ALFWorld gains are largest for Pick2 (+8.12 pp) and Pick (+7.98 pp),
while Cool decreases by 3.04 pp. The WebShop revisions solve 41 and 40
additional episodes out of 384. These gains measure guidance at fixed weights, not no-skill learning; successive WebShop revisions show nonmonotonic gains.

\begin{table}[!htbp]
\centering\normalfont\fontsize{8}{9.2}\selectfont
\setlength{\tabcolsep}{2pt}
\renewcommand{\arraystretch}{1.04}
\setlength{\heavyrulewidth}{.65pt}
\setlength{\lightrulewidth}{.4pt}
\setlength{\cmidrulewidth}{.4pt}
\definecolor{TableHeader}{HTML}{F0F2F5}
\definecolor{TableStudent}{HTML}{EAF0F8}
\colorlet{TableOursStudent}{TableStudent}
\colorlet{TableOursTeacher}{TableStudent}
\definecolor{TableGainBand}{HTML}{F4F5F6}
\colorlet{TableRoleStudent}{black}
\colorlet{TableRoleTeacher}{black}
\colorlet{TableGain}{black}
\colorlet{TableLoss}{black}
\colorlet{TableInterval}{black}
\renewcommand{\bestcell}[1]{\textbf{#1}}
\renewcommand{\tablefocus}[2]{{\setlength{\fboxsep}{.5pt}\setlength{\fboxrule}{.35pt}\fcolorbox{black}{TableStudent}{#2}}}
\renewcommand{\tablecite}[1]{{\fontsize{6.2}{7.2}\selectfont\citep{#1}}}
\renewcommand{\compacttablecite}[1]{\tablecite{#1}}
\caption{Skill-bank revision with a frozen policy.
Revision indexes bank updates with unchanged model weights.
Rates are percentages; $\Delta$ is the change from the first record (pp).
Initial guidance settings and comparison scope are discussed in
Appendix~\ref{app:ablation-records}.}
\label{tab:frozen-policy}
\begin{tabularx}{\linewidth}{c*{8}{>{\raggedleft\arraybackslash}X}}
\toprule
\multicolumn{9}{l}{\textbf{(a) ALFWorld: frozen Qwen3-1.7B}}\\[2pt]
\rowcolor{TableHeader}
\textbf{Revision} & \textbf{Pick} & \textbf{Look} & \textbf{Clean} & \textbf{Heat} & \textbf{Cool} & \textbf{Pick2} & \textbf{Macro SR} & $\boldsymbol{\Delta}$\\
\midrule
\rowcolor{TableStudent}
1 & 32.14 & 43.81 & 60.47 & 70.95 & 61.03 & 28.16 & 49.43 & --\\
\rowcolor{TableStudent}
2 & 37.84 & 44.71 & 61.94 & 72.38 & 57.77 & 34.06 & 51.45 & \textcolor{TableGain}{+2.02}\\
\rowcolor{TableStudent}
3 & 40.12 & 44.93 & 63.25 & 72.10 & 57.99 & 36.28 & \bestcell{52.45} & \textcolor{TableGain}{+3.02}\\
\bottomrule
\end{tabularx}
\vspace{2pt}
\begin{tabularx}{\linewidth}{c*{4}{>{\raggedleft\arraybackslash}X}c>{\raggedleft\arraybackslash}X}
\toprule
\multicolumn{7}{l}{\textbf{(b) WebShop: frozen Qwen2.5-3B}}\\[2pt]
\rowcolor{TableHeader}
\textbf{Revision} & \textbf{Seed 1} & \textbf{Seed 2} & \textbf{Seed 3} & \textbf{Mean SR} & \shortstack{\textbf{Successes /}\\\textbf{episodes}} & $\boldsymbol{\Delta}$\\
\midrule
\rowcolor{TableStudent}
0 & 58.59 & 59.38 & 54.69 & 57.55 & 221 / 384 & --\\
\rowcolor{TableStudent}
1 & 71.09 & 66.41 & 67.19 & \bestcell{68.23} & 262 / 384 & \textcolor{TableGain}{+10.68}\\
\rowcolor{TableStudent}
2 & 69.53 & 69.53 & 64.84 & 67.97 & 261 / 384 & \textcolor{TableGain}{+10.42}\\
\bottomrule
\end{tabularx}
\end{table}

\label{sec:dynamics}
In Figure~\ref{fig:learning-dynamics}(a,b), Student SR rises from
48.34\% to 55.72\% on ALFWorld (Cycles~1--4) and 69.53\% to 74.74\%
on WebShop (Cycles~1--3). Later WebShop records plateau near 74.74\%.
Shaded regions add training for \method{} only; comparisons use Cycle~3.

\subsection{Verification of Skill Revisions}
\label{sec:edit-verification}
Table~\ref{tab:edit-decisions} evaluates verification in
Figure~\ref{fig:cycle}(d). Each operation is compared with its removal
under a frozen 3B policy on matched task--seed pairs. Gains and losses
count failures changed to successes and the reverse. Acceptance requires
positive total net gain and no affected-family net regression.

\begin{table}[!htbp]
\centering\footnotesize
\setlength{\tabcolsep}{4pt}
\renewcommand{\arraystretch}{1.0}
\caption{ALFWorld-3B Cycle~3 operation verification at fixed weights.
Matched-pair counts compare each operation's inclusion with its removal.
Net is Gains minus Losses; accepted edits are highlighted.}
\label{tab:edit-decisions}
\begin{tabularx}{\linewidth}{@{}Xrrrrl@{}}
\toprule
\rowcolor{TableHeader}
\textbf{Operation} & \textbf{Pairs} & \textbf{Gains} & \textbf{Losses} & \textbf{Net} & \textbf{Decision}\\
\midrule
Add simple-search recovery & 48 & 1 & 1 & 0 & Reject\\
Rewrite clean readiness & 48 & 0 & 1 & \textcolor{TableLoss}{$-1$} & Reject\\
Rewrite cool-search frontier & 48 & 3 & 6 & \textcolor{TableLoss}{$-3$} & Reject\\
Rewrite heat visible acquisition & 48 & 0 & 1 & \textcolor{TableLoss}{$-1$} & Reject\\
Rewrite heat retention & 48 & 0 & 4 & \textcolor{TableLoss}{$-4$} & Reject\\
Rewrite invalid-action contract & 240 & 4 & 5 & \textcolor{TableLoss}{$-1$} & Reject\\
\rowcolor{TableGainBand}
Rewrite look-repeat recovery & 48 & 7 & 5 & \textcolor{TableGain}{\textbf{+2}} & \textcolor{TableGain}{\textbf{Accept}}\\
Rewrite Pick2 remaining acquisition & 48 & 3 & 7 & \textcolor{TableLoss}{$-4$} & Reject\\
\rowcolor{TableGainBand}
Rewrite Pick2 held-object placement & 48 & 7 & 3 & \textcolor{TableGain}{\textbf{+4}} & \textcolor{TableGain}{\textbf{Accept}}\\
Rewrite Pick2 search recovery & 48 & 3 & 6 & \textcolor{TableLoss}{$-3$} & Reject\\
\bottomrule
\end{tabularx}
\end{table}

Across three evaluation seeds, two of ten proposals pass, with pooled net
gains of two and four successes.
The cool-search rewrite is rejected despite three gains because it causes
six losses. The invalid-action rewrite improves Look by one success but
reduces Cool by two, motivating family-level guards. Verification filters
measured regressions before edits become persistent guidance, although
accepted edits can still harm individual trajectories. These conditional
effects do not sum to whole-bank gains (Appendix~\ref{app:edit-decisions});
Table~\ref{tab:ablation-core} separately evaluates how bank revision affects subsequent no-skill Student learning.
Together, these comparisons evaluate the framework at complementary levels:
revision-schedule controls measure learned Student performance, while matched
rule removals identify beneficial and harmful changes in the guidance itself.
Figure~\ref{fig:skill-case-study} presents a rule-retirement case with the
review prompt, trajectory evidence, and recorded rule-state change.

\subsection{Skill Routing on API Models}
\label{sec:api-results}
\begin{figure}[!htbp]
\centering
\includegraphics[width=.94\linewidth,trim=6.50bp 5.81bp 6.27bp 5.58bp,clip]{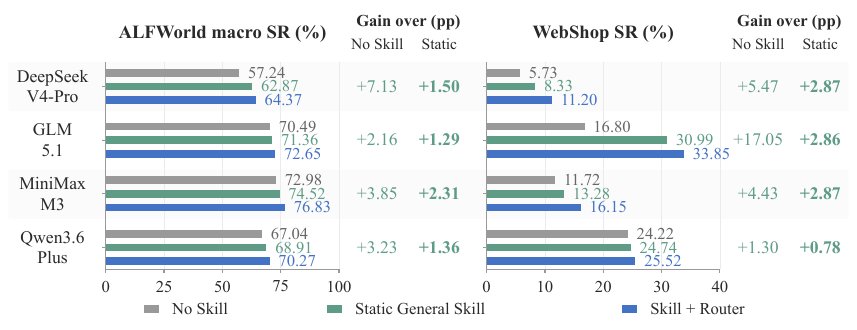}
\par\nointerlineskip
\caption{Skill routing across four API models.
Bars show three-seed SR; columns give gains over No Skill and Static (pp).
Axes start at zero with different ranges. Full results: Table~\ref{tab:api-results}.}
\label{fig:api-models}
\end{figure}
Figure~\ref{fig:api-models} tests external guidance on four API models
without parameter updates, using three evaluation seeds.
Skill + Router improves over No Skill by 2.16--7.13 pp on ALFWorld
macro SR and 1.30--17.05 pp on WebShop SR. Routing adds
1.29--2.31 pp and 0.78--2.87 pp, respectively, over Static General Skill.
Both aggregate metrics improve over static guidance for all four models;
Table~\ref{tab:api-results} gives family-level and pooled results.
The largest gains over No Skill occur for DeepSeek V4-Pro on ALFWorld
(+7.13 pp) and GLM 5.1 on WebShop (+17.05 pp), showing that the benefit
depends on both the model and environment. More consistently, routing
improves over static guidance in all eight model--benchmark comparisons.
This study isolates inference-time guidance without parameter updates;
the Student ablation separately establishes the gains from training with
an iteratively revised bank. Neither comparison implies that every task
family benefits equally.
\section{Conclusion}
We present \method{}, a framework that co-evolves agent policies and skill
banks through skill internalization and verified revision. Distillation
feedback serves two complementary purposes: improving the policy and
identifying skills that need revision. State-aware routing supplies
relevant guidance, while environment verification selects bank updates
for subsequent learning cycles. Experiments on ALFWorld and WebShop show
that continual revision improves skill-free performance over static banks
and a single update. Without inference-time skills, the 3B model achieves
$74.74\%$ success and an $88.37$ score on WebShop, yielding relative gains
of $0.73\%$ and $3.96\%$ over SOTA. These findings support a reciprocal
learning process in which policies internalize useful guidance and their
feedback helps maintain the skills used in subsequent training.
\section*{Limitations and Future Work}
Our study leaves open how the benefits of skill guidance translate into
lasting policy capabilities and when revision is most useful. Future work
could combine causal analysis with learning dynamics to guide adaptive
revision schedules. We also focus on individual rules and their
applicability; skill composition, hierarchical organization, and transfer
across tasks remain to be studied. Extending \method{} to these settings
could improve knowledge reuse and retention in lifelong learning.
\label{sec:main-end}
\clearpage
\subsection*{AI Use Statement}
Generative AI tools were used to assist with manuscript preparation and
language polishing. The authors take full responsibility for the content
and accuracy of the final manuscript.

\subsection*{Ethics Statement}
This work evaluates language agents in household simulation and benchmark
shopping environments; no human-subject study is reported. Use and
redistribution of benchmark assets and pretrained models remain subject to
their respective licenses and access conditions. Skills extracted from model
outputs and interaction histories can preserve errors or biases, and improved
task success does not establish safe behavior outside the evaluated settings.
The verification gates test task performance under a fixed policy, rather
than certify general safety. Physical or live-commerce deployment requires further safeguards for authorization, privacy, and unintended actions.

\subsection*{Reproducibility Statement}
Section~\ref{sec:method} defines the problem and method;
Appendices~\ref{app:interaction}--\ref{app:verification} detail interaction,
initialization, training, pseudocode, and verification.
Appendix~\ref{app:controlled-protocol} specifies the controlled protocol;
Appendices~\ref{app:result-protocol}--\ref{app:ablation-records} provide
evaluation and detailed results. We will release training code, executable
configurations, skill/router versions, task manifests, training and decoding
seed records, acceptance settings, episode-level outcomes, and aggregation
scripts, linking results to their checkpoint and protocol identities.

\clearpage
\bibliographystyle{iclr2027_conference}
\bibliography{references}

\clearpage
\appendix
\begin{center}
\Large\bfseries Appendix
\end{center}
\setcounter{topnumber}{4}
\setcounter{bottomnumber}{3}
\setcounter{totalnumber}{5}
\section{Related Work}
\label{sec:related}
\label{app:related-work}

\paragraph{External skills and agent memory.}
Temporally extended skills have a long history in reinforcement learning:
the options framework models them as closed-loop policies with initiation
and termination conditions \citep{sutton1999options}. Language agents also
reuse external artifacts. Toolformer learns when and how to invoke tools
through self-supervised training \citep{schick2023toolformer}, while
Voyager builds a library of
executable programs and improves them through environment feedback and
iterative prompting, without fine-tuning the underlying language model
\citep{wang2023voyager}. Agent Workflow Memory induces reusable workflows
from interaction experience \citep{wang2024awm}.
\method{} studies a coupled setting in which the policy also changes:
a skill bank supplies training supervision, and evidence from the updated
policy informs later revisions. Unlike executable policies, its procedural rules condition action
generation, and their utility depends on the policy.

\paragraph{Memory and context adaptation.}
Generative Agents combine experience retrieval with higher-level reflections
to support subsequent behavior \citep{park2023generative}. Reflexion retains
linguistic feedback across attempts \citep{shinn2023reflexion}, and ExpeL
extracts reusable insights from experience \citep{zhao2023expel}.
A-Mem dynamically organizes memories through indexing and linking
\citep{xu2025amem}; Agentic Context Engineering maintains evolving playbooks
through generation, reflection, and incremental curation \citep{zhang2025ace}.
These approaches motivate maintaining context as experience accumulates.
Our focus is its changing utility during policy learning: the bank provides
training-time guidance, and the updated policy supplies evidence for its
next revision. Environment acceptance gates decide which candidate changes
become persistent supervision, linking context maintenance to the requirements
of the evolving policy.

\paragraph{Reinforcement learning and self-distillation.}
GRPO estimates relative advantages from groups of sampled responses
\citep{shao2024grpo}; GiGPO extends group-based optimization to interactive
agents through episode-level and anchor-state groups \citep{feng2025gigpo}.
These methods provide outcome-based learning signals. Knowledge distillation
transfers teacher predictions to a student \citep{hinton2015distilling};
context distillation transfers behavior induced by additional context to
a model that does not receive that context \citep{snell2022context}.
On-policy distillation supplies teacher feedback on student-generated
sequences, reducing the mismatch between training examples and the
student's own generations \citep{agarwal2024onpolicy}.
Complementary self-distillation methods use contextual asymmetry to obtain
token-level supervision from the policy itself. OPSD conditions a same-model teacher
on privileged information while the student receives the original input
\citep{zhao2026opsd}. RLSD uses teacher--student differences to modulate
update magnitudes while retaining outcome-derived update directions
\citep{yang2026rlsd}. SDAR instead regulates an auxiliary distillation
objective with a bounded gate \citep{lu2026sdar}. \method{} builds on the
same-model, different-context construction; it additionally maintains the
procedural context as the policy learns. SDFT also uses a demonstration-conditioned
teacher for on-policy self-distillation, studying continual acquisition and
retention \citep{shenfeld2026sdft}. Our experiments instead examine the
interaction between skill internalization and persistent bank revision.

\paragraph{Skill internalization and utilization.}
SkillRL recursively evolves a hierarchical skill bank alongside policy
optimization \citep{xia2026skillrl}. Skill1 jointly trains skill selection,
utilization, and distillation under a shared task-outcome objective
\citep{shi2026skill1}. These approaches establish policy--library co-evolution
as a relevant setting. Our focus is the feedback produced during no-skill
Student learning: it localizes decisions for rule inspection, followed by
fixed-policy verification before revised guidance supervises another segment.
Skill0 gradually withdraws skill context during RL and evaluates its
on-policy helpfulness, aiming for autonomous inference without retrieval
\citep{lu2026skill0}. Skill0.5 combines general-skill internalization with
task-specific skill utilization through difficulty-aware routing
\citep{zhu2026skill05}. Skill-SD summarizes trajectories into skills that
condition a training-time teacher, transferring their guidance through
self-distillation \citep{wang2026skillsd}. These studies motivate measuring
both the no-skill policy and the benefit of retained external guidance.
\method{} fixes the bank within each learning segment and then revisits its
rules with the resulting policy held fixed. Its evaluation consequently
separates student improvement across segments from the contextual benefit
of the bank at a given checkpoint.

\paragraph{On-policy skills and verified revision.}
Self-Refine iteratively revises outputs using model-generated feedback
\citep{madaan2023selfrefine}. ProTeGi combines textual critiques with prompt
search \citep{pryzant2023protegi}, and TextGrad propagates language feedback
to optimize system components \citep{yuksekgonul2024textgrad}.
GEPA uses trajectory reflection and evaluation to evolve prompts
\citep{agrawal2026gepa}; Meta Agent Search explores code-defined agent
designs using an archive of prior discoveries \citep{hu2025adas}.
These methods optimize reusable agent components. SkillCycle revises the
procedural context that supplies supervision to a changing Student, making
the quality of an edit conditional on the current policy checkpoint.
Language Agent Tree Search combines tree search, reflection, and environment
feedback to improve action selection \citep{zhou2024lats}.
In \method{}, feedback instead evaluates edits to a persistent skill bank:
better wording alone does not establish that an edit improves the fixed
policy's behavior in the environment.
OPID provides the underlying supervision mechanism: it extracts hierarchical
hindsight skills from on-policy trajectories, routes critical-step guidance,
and rescores the same response with and without skills to construct a dense
training signal \citep{yang2026opid}. \method{} adopts this supervision
principle and investigates persistent, revisable guidance across training
segments. Candidate rule changes are screened using policy evidence and
subjected to paired environment comparisons before bank commitment;
rejected changes preserve the previous bank. This separates the role of
skills as distillation context from the decision to retain a particular
revision. The resulting research question is whether internalization feedback
can improve persistent guidance and, through it, subsequent no-skill learning.
UCOB couples mixed skill/no-skill rollouts with bidirectional distillation,
using anchor-state returns to choose the local teaching direction; its
credit signals also update skill utilities and train a reflection-based
skill writer \citep{tu2026ucob}. Both methods connect policy learning with
skill evolution. \method{} instead collects no-skill Student rollouts and
treats revision as an explicit intervention on a persistent rule bank:
after each learning segment, it fixes the policy, screens rule edits, and
requires targeted and whole-bank environment acceptance before commitment.
This separates the decision to change supervision from the policy updates
that subsequently use it. Our controlled comparisons evaluate this revision
schedule, while the main benchmark table compares reported performance
with RL and distillation methods using the same model families.
Our evaluations separate complementary comparisons. Cycle~0 is the
outcome-only initialization before \method{} training; continued GiGPO
provides the outcome-only RL comparison at the training endpoint. Static,
Refresh once, and \method{} compare fixed, one-time, and iterative bank
updates during skill-guided learning. The frozen-policy study separately
examines changes in guidance utility. Together, these comparisons distinguish
initialization, bank revision, and downstream Student learning
(Section~\ref{sec:ablations}).
\Needspace{6\baselineskip}
\section{Notation}
\label{app:notation}
\begingroup
\small
\renewcommand{\LTcaptype}{}
\setlength{\LTpre}{2pt}
\setlength{\LTpost}{4pt}
\begin{longtable}{@{}lp{0.70\linewidth}@{}}
\toprule
Symbol & Meaning\\
\midrule
\endfirsthead
\toprule
Symbol & Meaning (continued)\\
\midrule
\endhead
\bottomrule
\endfoot
\bottomrule
\endlastfoot
$x,\mathcal D$ & Task instruction and task distribution.\\
$s_t,o_t$ & Latent environment state and exposed observation.\\
$h_t,z_t$ & Observable interaction history and constructed model context.\\
$C,\mathcal C$ & Number of training segments and the context constructor.\\
$y_t,y_{t,j}$ & Generated response and its token at position $j$.\\
$a_t,\psi_x$ & Executed action and response-to-action interface.\\
$B,k_t$ & External skill bank and the selected document (or abstention).\\
$R$ & Fixed routing algorithm with the current declarative metadata.\\
$H_c=(B_c,R_c)$ & Committed bank/router state after revision $c$.\\
$N_c,\omega$ & Updates in segment $c$ and declared protocols/budgets.\\
$\theta_{c,u},M_c$ & Scoring snapshot within segment $c$ and its final student.\\
$\pi^S,\pi^T$ & Same-weight no-skill and skill-conditioned policy views.\\
$g_{t,j}$ & Signed log-probability difference on a sampled token.\\
$E_c$ & Training evidence passed to skill revision.\\
$U_x,J_{\mathcal D}$ & Specified trajectory metric and its expectation.\\
$q,\widehat\Delta(q)$ & Proposed rule edit and its paired estimated effect.\\
$L_c(t;k),d_{t,r}$ & Frozen-scorer action log likelihood and rule-removal influence.\\
$\Gamma^{\mathrm{tar}}_{\omega},\Gamma^{\mathrm{bank}}_{\omega}$ & Targeted and whole-bank acceptance gates on paired outcomes.\\
\end{longtable}
\endgroup

\section{Interaction Protocol}
\label{app:interaction}

We consider instruction-following tasks in partially observable environments,
as in skill-based language-agent learning
\citep{lu2026skill0,zhu2026skill05}. A task is specified by an instruction $x\sim\mathcal{D}$ and an
associated environment initialized from $\rho_x(s_1)$. At interaction
step $t$, the environment has a latent state $s_t$, exposes an observation
$o_t$, and accepts an action $a_t$. Its transition and observation kernels
are $P_x(s_{t+1}\mid s_t,a_t)$ and
$\Omega_x(o_{t+1}\mid s_{t+1},a_t)$, respectively. The agent has no direct
access to $s_t$. An episode ends at environment termination or a fixed
interaction budget $T_{\max}$.

Let $\mathcal{A}^{\mathrm{obs}}_t$ denote the action interface exposed to the
agent, including admissible commands or action templates. The information
available before acting is
\begin{equation}
 h_t=(x,o_1,y_1,a_1,\ldots,o_{t-1},y_{t-1},a_{t-1},o_t,
       \mathcal{A}^{\mathrm{obs}}_t),
 \label{eq:history}
\end{equation}
where $y_i$ is the agent's generated response at an earlier step. A fixed
context constructor $\mathcal C$ produces the model input $z_t=\mathcal C(h_t)$, possibly
retaining only a bounded history. The policy generates a response
$y_t=(y_{t,1},\ldots,y_{t,L_t})$ over a vocabulary $\mathcal{V}$:
\begin{equation}
 p_\theta(y_t\mid z_t)
 =\prod_{j=1}^{L_t}\pi_\theta(y_{t,j}\mid z_t,y_{t,<j}),
 \qquad a_t=\psi_x(y_t,\mathcal{A}^{\mathrm{obs}}_t).
 \label{eq:response-action}
\end{equation}
Here $\psi_x$ is the environment-specific action parser and execution
interface, including its prescribed invalid-output handling. The response may
contain reasoning or formatting tokens in addition to the action; only the
parsed action affects the environment. Throughout, $t$ indexes environment interactions, whereas $j$ indexes tokens within a generated response.

\subsection{Contextual Policy Views and Performance}
The router selects at most one skill document,
$k_t=R(h_t;B)\in B\cup\{\varnothing\}$. The two policy views in
Section~\ref{sec:problem} are defined over the same vocabulary and parameters:
\begin{equation}
\begin{aligned}
 \pi_\theta^{S}(v\mid h_t,y_{t,<j})
   &=\pi_\theta(v\mid \mathcal C(h_t),y_{t,<j}),\\
 \pi_\theta^{T}(v\mid h_t,k_t,y_{t,<j})
   &=\pi_\theta(v\mid \mathcal C(h_t;k_t),y_{t,<j}),
 \qquad v\in\mathcal V.
\end{aligned}
\label{eq:contextual-views}
\end{equation}
Here $\mathcal C(h_t;k_t)$ inserts the selected skill and
$\mathcal C(h_t;\varnothing)=\mathcal C(h_t)$. The Teacher's additional
context is procedural guidance; its predictions need not be better than
the Student's on every decision.

For trajectory metric $U_x$ (success or task score), let
\begin{equation}
 J_{\mathcal D}(\theta;B,R)
 =\mathbb E_{x\sim\mathcal D}\,
  \mathbb E_{\tau\sim p_{\theta,B,R}(\cdot\mid x)}[U_x(\tau)],
 \qquad
 J_{\mathcal D}(\theta;\varnothing)
 :=J_{\mathcal D}(\theta;\varnothing,R_{\varnothing}),
 \label{eq:performance}
\end{equation}
where $p_{\theta,B,R}$ is the induced trajectory distribution and
$R_{\varnothing}$ always abstains. SkillCycle targets
$J_{\mathcal D}(\theta_C;\varnothing)$; skill-conditioned evaluation measures the remaining context benefit at fixed weights.

\section{Policy Initialization}
\label{app:initialization}
\label{sec:gigpo-pretraining}
Cycle~0 denotes the outcome-only base policy before \method{} OPD updates.
Subsequent cycle indices count skill-guided learning segments; the
frozen-policy study instead indexes bank revisions. The standardized
initialization protocol uses GiGPO
\citep{feng2025gigpo}. For a prompt group
$i$, let $G_{i,k}$ be the episode return of rollout $k$. The episode-relative
advantage is
\begin{equation}
 A^{\mathrm{episode}}_{i,k}
 =\operatorname{Norm}\!\left(G_{i,k};\{G_{i,k'}\}_{k'=1}^{K}\right),
\end{equation}
where $\operatorname{Norm}$ centers the return within the group and, if
selected by the protocol, scales it by the group standard deviation.
Within a prompt group, matching anchor observations define state groups. Normalizing step returns within each group yields $A^{\mathrm{step}}_t$, giving
\begin{equation}
 A_t=A^{\mathrm{episode}}_t+w_{\mathrm{step}}A^{\mathrm{step}}_t,
 \label{eq:gigpo-advantage}
\end{equation}
and is broadcast over valid response tokens for the PPO actor update.
The current WebShop configuration uses $K=8$, $w_{\mathrm{step}}=1.0$,
and mean-centering (\texttt{mean\_norm}), explicitly
$\operatorname{Norm}(v_i;\{v_j\}_{j=1}^{K})=v_i-K^{-1}\sum_jv_j$.
Anchor states are matched exactly in this configuration. ALFWorld uses benchmark-specific normalization and matching. Neither initializer uses skills.

The reference initialization configuration specifies 90 training updates,
saves a checkpoint every two updates, retains four, and merges
\texttt{global\_step\_90}. The WebShop starting checkpoint was saved at
step 90 of a nominal 150-update run. The benchmark starting checkpoints are listed below.

\paragraph{Benchmark starting points.}
The 1.7B ALFWorld Cycle~0 student is the locally evaluated
\texttt{Jinyang23/OPID-ALFWorld-1.7B} checkpoint; the reported macro SR is
45.89\%. The 1.7B WebShop Cycle~0 student is a clean GiGPO step-90 checkpoint.
The 7B reference uses the public GiGPO checkpoints for ALFWorld and WebShop,
with recorded revisions \texttt{895c9fe1} and \texttt{8e58a29c}, respectively.
The 3B benchmark trajectory begins at Cycle~1; its reported gain is measured
from that checkpoint. Cross-model trajectories describe learning from their
respective starting points. Matched initializations apply within each
controlled ablation (Appendix~\ref{app:controlled-protocol}).
The WebShop initialization configuration uses a learning rate of
$10^{-6}$; initialization and the subsequent eight-update Phase~1 segments
are counted separately.

\section{Training Procedure and Implementation Details}
\label{app:training}
\label{app:implementation-scope}
\subsection{Differentiable Objective and Token Reductions}
\label{app:objective}
During an actor update, write $\bar\theta=\theta_{c,u}$ for the frozen
rollout/scoring snapshot and $\theta$ for the parameters being optimized.
The current-weight Teacher uses $\bar\theta$ with skill context; the old
Student uses the same snapshot without that context. Both scores in
Eq.~\eqref{eq:opd-gap} are detached. Thus the auxiliary OPD gradient is
$-\mathcal N_{m^o}(g\nabla_\theta\ell(\theta))$: a positive gap increases
the sampled token's probability and a negative gap decreases it.
This signed sampled-token loss is not a full-vocabulary KL divergence.
The next rollout and Teacher rescore use the updated Student snapshot.
The verified alternative response $z_t$ has no-skill Student log probability
$\ell^z_{t,j}(\theta)=\log\pi^S_\theta(z_{t,j}\mid h_t,z_{t,<j})$.

Let $d_t$ identify appended demonstration rows. The masks are
\begin{equation}
 m^{\mathrm{RL}}=m^{\mathrm{resp}}(1-d_t),\qquad
 m^o=m^{\mathrm{RL}}m^{\mathrm{applied}},\qquad
 m^d=d_t m^{\mathrm{supervision}}.
 \label{eq:training-masks}
\end{equation}
Prompt, observation, skill-context, and padding tokens receive no direct
loss. Demonstration rows have zero OPD-applied masks and are excluded from
PPO, entropy, and reference regularization. In the WebShop 1.7B configuration,
$m^{\mathrm{supervision}}$ selects only the alternative action span; the WebShop 3B configuration selects all valid tokens in the verified alternative response.

For microbatch $b$, define
$\langle v\rangle_{m,b}=\sum_{(t,j)\in b}m_{t,j}v_{t,j}/n_b$, with
$n_b=\sum_{(t,j)\in b}m_{t,j}$ and zero for an empty mask. For $M$
microbatches in one local actor minibatch, the two reductions are
\begin{equation}
 \mathfrak R_m(v)=\frac1M\sum_b\langle v\rangle_{m,b},\qquad
 \mathfrak T_m(v)=\sum_b\frac{n_b}{N}\langle v\rangle_{m,b},
 \quad N=\sum_b n_b.
 \label{eq:loss-reduction}
\end{equation}
Set $\mathfrak T_m=0$ when $N=0$. PPO and its regularizers use
$\mathfrak R$. WebShop 1.7B auxiliary losses use $\mathcal N=\mathfrak T$:
each microbatch auxiliary loss is scaled by $Mn_b/N$ before the $1/M$
gradient-accumulation factor. This prevents sparse demo supervision from
depending on its distribution across microbatches, including a smaller
final minibatch. WebShop 3B demonstration supervision uses
$\mathcal N=\mathfrak R$. Distributed workers average gradients; they do
not reweight them by a global active-token count.

Let $\rho=\exp(\ell(\theta)-\ell(\bar\theta))$. The implemented
dual-clipped PPO loss is
\begin{equation}
\begin{aligned}
 b(\rho,A)&=\max\{-\rho A,-\operatorname{clip}(\rho,0.8,1.2)A\},\\
 f(\rho,A)&=\begin{cases}\min\{b(\rho,A),-3A\},&A<0,\\
 b(\rho,A),&A\geq0,\end{cases}\qquad
 \mathcal L_{\mathrm{PPO}}(A)=\mathfrak R_{m^{\mathrm{RL}}}(f(\rho,A)).
\end{aligned}
\label{eq:explicit-ppo}
\end{equation}
With the segment-frozen reference log ratio $v=\log\pi_{\mathrm{ref}}-\ell(\theta)$, the regularizers are
\begin{equation}
 \mathcal L_{\mathrm{ref}}=\mathfrak R_{m^{\mathrm{RL}}}
    (\operatorname{clip}(e^v-v-1,-10,10)),\qquad
 \mathcal H=\mathfrak R_{m^{\mathrm{RL}}}
    \left(-\sum_a\pi^S_\theta(a)\log\pi^S_\theta(a)\right).
 \label{eq:actor-regularizers}
\end{equation}
The reference policy is distinct from the current-weight contextual
Teacher. Actor parameters are updated with AdamW on
Eq.~\eqref{eq:webshop-objective}, carrying its optimizer state:
\begin{equation}
 (\theta^+,s^+)=\operatorname{AdamW}
    (\theta,s,\nabla_\theta\mathcal L;\mathrm{lr},\mathrm{wd}).
 \label{eq:policy-update}
\end{equation}
The actor optimizer defaults are one PPO epoch, weight decay $0.01$,
and gradient-norm clipping at 1. These are distinct from PPO ratio
clipping and gap clipping. Auxiliary OPD uses the raw detached gap
($\texttt{delta}$ weighting, no clipping); $c_g$ affects only the
advantage injection, which is disabled at $\eta=0$.

\subsection{Environment Advantages and Verified Alternatives}
\label{sec:benchmark-supervision}
\paragraph{WebShop hierarchical feedback.}
In the WebShop 1.7B configuration, each prompt group contains eight
distinct trajectories. For each trajectory $i$, take the maximum recorded
success, score, and progress values, giving $v_i^{(k)}$ for $k=1,2,3$.
Choose the first coordinate in that order whose within-group range
exceeds $10^{-6}$, and compute the centered environment advantage
\begin{equation}
 A_i^{\mathrm{env}}=s_k\left(v_i^{(k)}-\frac18\sum_{i'=1}^8v_{i'}^{(k)}\right),
 \qquad (s_1,s_2,s_3)=(1,1.8,1.8).
 \label{eq:web-hierarchical-advantage}
\end{equation}
If all coordinates are constant, the advantage is zero. Otherwise it is
broadcast to the trajectory's valid response tokens, without standard
deviation normalization. This prioritizes success whenever it separates
rollouts, using denser signals only when the preceding coordinates do not.
Rollouts have distinct IDs and sampling-request seeds; exclude batches if all groups contain identical responses.

\paragraph{ALFWorld native grouping.}
The 3B Cycle~3 configuration uses episode mean normalization and zero
state-group advantage weight. Let
$r_i$ be the sum of token rewards in response row $i$ and
$\mathcal G_i$ contain rows with its prompt identity. The environment advantage is
\begin{equation}
 A_i^{\mathrm{env}}=r_i-b_i,\qquad
 b_i=\begin{cases}|\mathcal G_i|^{-1}\sum_{j\in\mathcal G_i}r_j,
                   &|\mathcal G_i|>1,\\0,&|\mathcal G_i|=1.
       \end{cases}
 \label{eq:alf-native-advantage}
\end{equation}
The default groups response rows across interaction steps, without
trajectory deduplication or standard deviation normalization; this is
not leave-one-out centering. The configured invalid-action penalty is
$0.01$. Gap clipping uses the 95th percentile of $|g|$ measured in a
calibration pass with actor updates disabled, then held fixed for the
eight formal updates.

\paragraph{WebShop 1.7B alternative-action verification.}
At a routed Student state, a valid original action with negative
action-mean gap is eligible for replacement. Unique trajectory--step
candidates are ordered by this gap, most negative first; the pool is 48
and at most eight alternatives are accepted per update. The proposed
action must be syntactically valid, admissible, and different from the
original. Each branch resets the same task and replays the same history
before executing one action. Branches are compared by success first,
then progress, using a progress margin greater than $10^{-8}$. A better
branch is accepted and a worse branch rejected. Only for tied outcomes
does the verifier require the two-sided condition
$\bar g_{\mathrm{original}}<0<\bar g_{\mathrm{alternative}}$.
Replay or execution errors cause rejection. A shortfall retains the
actual accepted count; no accepted action gives zero demo loss.
The demonstration loss supervises action tokens under the bare Student
prompt. A tied branch passing the probability check is preference
supervision, not evidence of greater environment success.

\subsection{Training Configurations}
\label{app:run-configurations}
Table~\ref{tab:audited-training-configs} specifies three configurations of
Eq.~\eqref{eq:webshop-objective}: auxiliary OPD with verified demonstrations,
verified demonstrations alone, and OPD advantage injection. Within each
bank-update ablation, the optimization configuration is held fixed and
only the revision schedule changes (Appendix~\ref{app:controlled-protocol}).

\begin{table}[htbp]
\centering\small
\caption{Cycle~3 optimization configurations.}
\label{tab:audited-training-configs}
\setlength{\tabcolsep}{4pt}
\begin{tabularx}{\linewidth}{@{}lXXX@{}}
\toprule
Setting & WebShop 1.7B & WebShop 3B & ALFWorld 3B\\
\midrule
$(\eta,\lambda_o,\lambda_d)$ & $(0,.001,.1)$ & $(0,0,.1)$ & $(.1,0,0)$\\
Demo tokens / reduction & Action / $\mathfrak T$ & Full response / $\mathfrak R$ & Disabled\\
Formal actor updates & 8 & 8 & 8\\
Learning rate & $10^{-6}$ & $10^{-6}$ & $10^{-6}$\\
Reference coefficient $\beta$ & $.01$ & $.01$ & $.01$\\
Configuration & Auxiliary + demo & Demo & Advantage injection\\
\bottomrule
\end{tabularx}
\end{table}

The WebShop 1.7B profile uses train batch 16, rollout group size 8,
PPO minibatch 64, microbatch 4 per GPU, two GPUs, maximum prompt/response
lengths 4096/512, horizon 15, history length 2, and entropy coefficient
$0.001$. WebShop 3B uses the same train batch, PPO minibatch, maximum
sequence lengths, horizon, and history length, with microbatch 8 per GPU
and four GPUs. Its OPD signal selects demonstration candidates and supports
rule diagnosis; parameter updates use PPO and verified demonstration
supervision without an auxiliary OPD term.

The ALFWorld 3B Cycle~3 configuration uses train batch 16, PPO minibatch
256, microbatch 16 per GPU, two GPUs, maximum lengths 2048/128,
horizon 30, history length 5, and validation batch 160. The release will
provide the per-result ALFWorld configuration assignments and checkpoints.

\paragraph{Teacher synchronization.}
The current-weight configuration in Eq.~\eqref{eq:opd-gap} synchronizes
Teacher and Student at each rollout update. The segment-frozen variant
holds Teacher parameters at $\theta_{c,0}$ throughout a segment.
The earlier ALFWorld 1.7B configuration uses this variant with $\eta=.001$,
gap clip 5, learning rate $2\times10^{-7}$, and reference coefficient $.05$.
Configuration manifests distinguish these synchronization schedules and
specify realized update counts for each reported result.

\paragraph{Supervision consistency.}
Supervision records retain response IDs, old Student and Teacher log
probabilities, signed gaps, and aligned response/applied/action masks.
They validate finite values, score-difference identities, and mask
containment. Routing records retain document/rule versions and selection
or abstention reasons. Repeated bindings are suppressed until observed
progress. The Teacher has no independent optimizer. The release will
link each result to its checkpoint, bank, complete resolved configuration,
and training/scoring synchronization schedule.

\paragraph{Evaluation configuration.}
The ALFWorld evaluation configuration fixes 128 tasks, decoding seeds
$\{1,2,3\}$, a 30-step horizon, a history length of five, and temperature
$0.4$. Evaluation checks episode completeness, requires active routes
in the skill-conditioned arm, and rejects routed records in the no-skill arm.
It computes success rates per seed and task family, keeping decoding
variation separate from variation across independently trained policies.
The WebShop GiGPO initializer specifies 90 updates, eight rollouts
per task, a learning rate of $10^{-6}$, $\gamma=0.95$, and
\texttt{mean\_norm} with $w_{\mathrm{step}}=1$. These settings describe
initialization, separately from the subsequent skill-guided segments.
Run-specific task manifests, acceptance configurations, independent-training
seed records, and interval-generation scripts will accompany the open-source
release to document each comparison.

\section{Complete Training Procedure}
\label{app:algorithm}
Algorithm~\ref{alg:skillcycle} gives the sequential-verification form of
Section~\ref{sec:method}. The ALFWorld operation records in
Appendix~\ref{app:numerical-gates} instead use candidate-removal controls
for the targeted test. Equation~\eqref{eq:cycle} in the main text defines the alternating updates
under training protocol and budgets $\omega$.
Segment $c$ contains $N_c$ updates, with $\theta_{c,0}=\theta_{c-1}$ and
$\theta_c=\theta_{c,N_c}$. The no-skill Student generates trajectories as follows:
\begin{equation}
 y_t\sim p^S_{\theta_{c,u}}(\cdot\mid h_t),\qquad
 k_t=R_{c-1}(h_t;B_{c-1}),\qquad
 a_t=\psi_x(y_t,\mathcal A_t^{\mathrm{obs}}).
 \label{eq:student-rollout}
\end{equation}
The router uses pre-action history. Teacher rescoring uses the same parameter
snapshot and response prefixes; no independent Teacher optimizer is used.

\begin{algorithm}[!ht]
\caption{Detailed SkillCycle training and verification}
\label{alg:skillcycle}
\centering
\begin{minipage}{0.98\linewidth}
\small\raggedright
\textbf{Input:} $\theta_0$, $H_0=(B_0,R_0)$, $\mathcal D_{\mathrm{train}}$,
$\mathcal D_{\mathrm{acc}}$, $C$, update counts $\{N_c\}$, protocol $\omega$.\par
\textbf{Output:} No-skill student $M_C=\pi_{\theta_C}$ and last committed $H_{C-1}$.\par
\vspace{4pt}
\newcommand{\skillstep}[3]{#1 & \hspace*{#2em}#3\\}
\renewcommand{\arraystretch}{1.07}
\begin{tabularx}{\linewidth}{@{}r@{\hspace{0.7em}}X@{}}
\skillstep{1}{0}{\textbf{for} $c=1,\ldots,C$ \textbf{do}}
\skillstep{2}{1}{$\theta_{c,0}\leftarrow\theta_{c-1}$; $E_c\leftarrow\varnothing$
 \hfill\emph{Phase 1: fix $H_{c-1}$}}
\skillstep{3}{1}{\textbf{for} $u=0,\ldots,N_c-1$ \textbf{do}}
\skillstep{4}{2}{$x_b\sim\mathcal D_{\mathrm{train}}$,
 $\tau_b\sim p_{\theta_{c,u},\varnothing}(\cdot\mid x_b)$;
 $\mathcal T_{c,u}\leftarrow\{\tau_b\}_b$}
\skillstep{5}{2}{$k_t\leftarrow R_{c-1}(h_t;B_{c-1})$ for each decision $t\in\mathcal T_{c,u}$}
\skillstep{6}{2}{$g_{t,j}\leftarrow\operatorname{sg}[\log\pi^T_{\theta_{c,u}}(y_{t,j}\mid h_t,k_t,y_{t,<j})
 -\log\pi^S_{\theta_{c,u}}(y_{t,j}\mid h_t,y_{t,<j})]$}
\skillstep{7}{2}{$A_{c,u}\leftarrow\operatorname{Adv}_{\omega}(\mathcal T_{c,u})$;
 $Z_{c,u}\leftarrow\operatorname{Aux}_{\omega}(\mathcal T_{c,u})$}
\skillstep{8}{2}{Optimize Eq.~\eqref{eq:webshop-objective} with the masks and
 reductions in Appendix~\ref{app:training}, obtaining $\theta_{c,u+1}$}
\skillstep{9}{2}{$E_c\leftarrow E_c\cup\operatorname{Log}
 (\mathcal T_{c,u},k,g,m,\theta_{c,u},H_{c-1})$}
\skillstep{10}{1}{\textbf{end for}}
\skillstep{11}{1}{$\theta_c\leftarrow\theta_{c,N_c}$;
 \textbf{if} $c=C$ \textbf{then break}}
\skillstep{12}{1}{$\widetilde E_c\leftarrow\operatorname{Select}_{\omega}(E_c)$
 \quad [Eqs.~\eqref{eq:information}--\eqref{eq:evidence-selection}]
 \hfill\emph{Phase 2: fix $\theta_c$}}
\skillstep{13}{1}{$d\leftarrow\operatorname{Influence}_{\theta_c}(\widetilde E_c)$
 \quad [Eq.~\eqref{eq:rule-influence}]}
\skillstep{14}{1}{$\mathcal Q\leftarrow\operatorname{Propose}_{\omega}
 (H_{c-1},\widetilde E_c,d)$}
\skillstep{15}{1}{$(q_1,\ldots,q_m)\leftarrow\operatorname{Screen}_{\omega}
 (\mathcal Q;\theta_c,\widetilde E_c)$; $m\leq K_{\omega}$}
\skillstep{16}{1}{$H^{(0)}\leftarrow H_{c-1}$}
\skillstep{17}{1}{\textbf{for} $i=1,\ldots,m$ \textbf{do}}
\skillstep{18}{2}{$H^{(i)}\leftarrow H^{(i-1)}$}
\skillstep{19}{2}{\textbf{if} $o(q_i)=\textsc{Keep}$ \textbf{then} log retention; \textbf{continue}}
\skillstep{20}{2}{\textbf{if} $\neg\operatorname{Valid}_{\omega}(q_i,H^{(i-1)})$
 \textbf{then} log rejection; \textbf{continue}}
\skillstep{21}{2}{$\widetilde H^{(i)}\leftarrow T_{q_i}(H^{(i-1)})$}
\skillstep{22}{2}{$V_i\leftarrow\mathcal V_c
 (\widetilde H^{(i)},H^{(i-1)};\mathcal P_{q_i})$}
\skillstep{23}{2}{\textbf{if} $\Gamma^{\mathrm{tar}}_{\omega}(V_i)=1$
 \textbf{then} $H^{(i)}\leftarrow\widetilde H^{(i)}$}
\skillstep{24}{2}{$\operatorname{Log}(q_i,V_i,H^{(i-1)},H^{(i)})$}
\skillstep{25}{1}{\textbf{end for}}
\skillstep{26}{1}{$V_{\mathrm{bank}}\leftarrow\mathcal V_c
 (H^{(m)},H_{c-1};\mathcal P_{\mathrm{bank}})$; $H_c\leftarrow H_{c-1}$}
\skillstep{27}{1}{\textbf{if} $\Gamma^{\mathrm{bank}}_{\omega}(V_{\mathrm{bank}})=1$
 \textbf{then} $H_c\leftarrow H^{(m)}$}
\skillstep{28}{1}{$\operatorname{Log}(V_{\mathrm{bank}},H_{c-1},H_c)$}
\skillstep{29}{0}{\textbf{end for}}
\skillstep{30}{0}{\textbf{return} $(\pi_{\theta_C},H_{C-1})$}
\end{tabularx}
\vspace{4pt}
$\operatorname{Adv}_{\omega}$ computes environment advantages;
$\operatorname{Aux}_{\omega}$ supplies verified auxiliary actions under the
current $(\theta_{c,u},H_{c-1})$, or $\varnothing$ when disabled.
In $\operatorname{Log}$, $m$ collects the response, applied, and action masks;
records retain scoring snapshots, bank and rule version identities, verification decisions, and interaction costs.\par
\end{minipage}
\end{algorithm}

\section{Routing, Screening, and Acceptance Details}
\label{app:method-details}
\subsection{Observable Critical-First Skill Routing}

\label{app:detail-routing}
\paragraph{Observable applicability.}
Write $B=B^G\mathbin{\dot\cup}B^K$ for general-workflow and critical-event
documents. Each document contains identifiable rules and declarative
applicability, blocking, priority, and provenance metadata. The fact
constructor $f_t=F(h_t)$ uses the instruction, observations, exposed actions,
and an episode-local record of confirmed progress and failed actions.
Progress is confirmed by post-action observations; the record resets each episode.

For document $b$, let $W_b$ and $V_b$ be required and blocking predicates,
and let $\beta_b(f_t)$ indicate valid bindings, exposed-action checks, and
repeat admissibility. Eligibility is
\begin{equation}
 e_b(f_t)=\mathbf 1[b\text{ active}]\,\beta_b(f_t)
 \prod_{w\in W_b}\mathbf 1[w(f_t)=\mathrm{true}]
 \prod_{v\in V_b}\mathbf 1[v(f_t)=\mathrm{false}].
 \label{eq:eligibility}
\end{equation}
Unknown facts cannot satisfy these conditions. Neither hidden states nor future observations enter $F$; priority ranks only documents whose eligibility checks pass.

\paragraph{Selection and abstention.}
Define $\mathcal K_t=\{b\in B^K:e_b(f_t)=1\}$ and
$\mathcal G_t=\{b\in B^G:e_b(f_t)=1\}$. For eligible set $\mathcal S$,
let $\mathcal A(\mathcal S,f)$ be the set of documents with maximal
lexicographic rank (declared priority, then predicate specificity), with
$\mathcal A(\varnothing,f)=\varnothing$. The router is
\begin{equation}
\begin{aligned}
 \mathcal S_t&=\begin{cases}
  \mathcal K_t,&\mathcal K_t\ne\varnothing,\\
  \mathcal G_t,&\mathcal K_t=\varnothing,
 \end{cases}
 &\qquad
 k_t&=\begin{cases}
  b,&\mathcal A(\mathcal S_t,f_t)=\{b\},\\
  \varnothing,&|\mathcal A(\mathcal S_t,f_t)|\ne1.
 \end{cases}
\end{aligned}
\label{eq:critical-router}
\end{equation}
A critical-document tie causes abstention; general fallback occurs only
when no critical document is eligible. Eligibility is recomputed at each
decision, and repeated bindings are suppressed until confirmed progress.
The router supplies document text, while the model chooses the action.
This design allows a critical-event instruction to interrupt a general
workflow only when its observable preconditions hold. Abstention avoids
injecting arbitrarily chosen guidance when those conditions leave an
unresolved tie. Routing and implementation details appear in Appendix~\ref{app:implementation-scope}.

\subsection{Evidence-Driven Rule and Candidate Screening}
\label{app:detail-screening}
Screening maps $E_c$ to a budgeted, ordered proposal list while holding
$\theta_c$ fixed. It uses training-side records; the proposer receives
neither acceptance-task contents nor final-test trajectories.

\paragraph{Selecting informative decisions.}
After checking token alignment, masks, finite scores, and the identity in
Eq.~\eqref{eq:opd-gap}, define the diagnostic mask and disagreement as
\begin{equation}
 m^a_{t,j}=m^{\mathrm{resp}}_{t,j}m^{\mathrm{applied}}_{t,j}
             m^{\mathrm{action}}_{t,j},\qquad
 n_t=\sum_jm^a_{t,j},\qquad
 I_t=\frac{1}{n_t}\sum_jm^a_{t,j}|g_{t,j}|,
 \label{eq:information}
\end{equation}
requiring $n_t>0$. This action-content mask is used for diagnosis, whereas
Eq.~\eqref{eq:policy-update} supervises the full valid response.
Observed feedback assigns
\begin{equation}
 \ell_t=\begin{cases}
 -1,&a_t\text{ is invalid},\\
 +1,&a_t\text{ is valid and terminates the episode successfully},\\
 \bot,&\text{otherwise}.
 \end{cases}
 \label{eq:action-label}
\end{equation}
Intermediate actions remain unknown even in successful episodes.
Let $E_c^L$ and $E_c^U$ denote valid labeled and unknown records. The selected
set is
\begin{equation}
 \widetilde E_c=E_c^L\cup\operatorname{Cap}_{\omega}
 \bigl(\operatorname{Tail}_{\gamma,\omega}(E_c^U)\bigr),
 \label{eq:evidence-selection}
\end{equation}
where $\operatorname{Tail}_{\gamma,\omega}$ selects high-disagreement records
within each training-seed/task-family group. The implementation sorts by
$(I_t,\mathrm{decision\_id})$ and retains the tail after the first
$\lfloor\gamma N\rfloor$ records, where $N$ is the group size and $\gamma$
defaults to $0.8$; this fixes tie handling at the boundary. Episode and task
caps limit unknown records; labeled records bypass them.
Disagreement prioritizes inspection; observed outcomes supply the direction
of evidence. A large gap alone can arise from either useful guidance or a
misleading rule, which is why it cannot serve as a positive action label.

\paragraph{Rule influence under a fixed scorer.}
Equation~\eqref{eq:rule-influence} defines the action-masked log likelihood
and rule-removal influence for a selected record.
Only exposed rules are compared. Both contexts are rendered separately and
scored with the same frozen $\theta_c$ and response; earlier cached scores
are not reused. The influence $d_{t,r}$ measures a likelihood change, not
task-return improvement. Sparse rescoring and its validation are described in
Appendix~\ref{app:screening}.

\paragraph{Proposals and likelihood screening.}
An offline proposer maps $(H_{c-1},\widetilde E_c,\{d_{t,r}\})$ to edits
$q$ with operation
$o(q)\in\{\textsc{Keep},\textsc{Rewrite},\textsc{Drop},\textsc{Add}\}$.
Each edit carries its target document and level, proposed text or metadata,
and supporting evidence; edits to existing rules also identify their versions.
An \textsc{Add} proposal may address a failed decision with no matching rule,
so it requires no parent-rule reference or rule-removal score.
For a text rewrite, let $k_t^{(q)}$ denote the edited context and define
\begin{equation}
\begin{aligned}
 \delta_t(q)&=L_c(t;k_t^{(q)})-L_c(t;k_t),\\
 S(q)&=\operatorname{HMean}_{t\in\widetilde E_c^L}
          [\ell_t\delta_t(q)],\qquad
 A(q)=\operatorname{HMean}_{t\in\widetilde E_c}|\delta_t(q)|.
\end{aligned}
 \label{eq:rewrite-screen}
\end{equation}
The hierarchical mean averages applicable records within tasks, then tasks
within seeds, then seeds. Rewrites with $A(q)\leq\varepsilon_\omega$ or
negative $S(q)$ are filtered out; $S(q)$ is used only when labels exist.
Unknown records provide magnitude information only. Ranking retains at most
$K_\omega$ proposals for confirmation (Appendix~\ref{app:screening}). These
likelihood filters apply to text rewrites; all mutations, including additions
and routing edits, require environment verification.

\subsection{Paired Verification and Skill Bank Update}
\label{app:detail-bank-update}
\paragraph{Rule-level interventions.}
Let $(q_1,\ldots,q_m)$ be the screened proposals in the declared order and
$H^{(0)}=H_{c-1}$ the working state. An edit operator
$T_q$ constructs a trial state from the current working state, subject to
format, reference/version, and conflict checks. \textsc{Keep} leaves the
state unchanged; \textsc{Rewrite}, \textsc{Drop}, and \textsc{Add} respectively
replace, deactivate, or insert a rule. Verification compares each trial
with the current state under the same frozen $\theta_c$.

For states $H'$ and $H$ and a nonempty matched panel $\mathcal P$, define
$\mathcal V_c(H',H;\mathcal P)$ as the paired outcomes
$\bigl(U_x(\tau^{H'}_{x,\xi}),U_x(\tau^H_{x,\xi})\bigr)_{(x,\xi)\in\mathcal P}$,
where both trajectories use $\theta_c$. For valid candidate $q_i$,
$\widetilde H^{(i)}=T_{q_i}(H^{(i-1)})$ yields the paired estimate in
Eq.~\eqref{eq:candidate-effect}.
The targeted panel $\mathcal P_{q_i}$ covers relevant development tasks,
with matched task/reset identities, seeds, decoding, and environment
interfaces. Rule deletion compares removal with retention; other edits
compare the trial with the current state.

\paragraph{Targeted acceptance and bank commitment.}
Let $\Gamma^{\mathrm{tar}}_{\omega}$ and
$\Gamma^{\mathrm{bank}}_{\omega}$ be binary gates on paired outcomes.
Their metrics, permissible regressions, uncertainty treatment, panels, and
budgets are declared before evaluation (Appendix~\ref{app:verification}).
For valid non-\textsc{Keep} edits, the sequential update is
\begin{equation}
 H^{(i)}=\begin{cases}
 \widetilde H^{(i)},&
 \Gamma^{\mathrm{tar}}_{\omega}\!\left(
 \mathcal V_c(\widetilde H^{(i)},H^{(i-1)};\mathcal P_{q_i})\right)=1,\\
 H^{(i-1)},&\text{otherwise}.
 \end{cases}
 \label{eq:targeted-update}
\end{equation}
\textsc{Keep} and invalid or conflicting proposals leave $H^{(i-1)}$
unchanged and receive distinct decision records. Acceptance is provisional; the whole-bank panel $\mathcal P_{\mathrm{bank}}$ from $\mathcal D_{\mathrm{acc}}$ determines
\begin{equation}
 H_c=\begin{cases}
 H^{(m)},&\Gamma^{\mathrm{bank}}_{\omega}\!\left(
 \mathcal V_c(H^{(m)},H_{c-1};\mathcal P_{\mathrm{bank}})\right)=1,\\
 H_{c-1},&\text{otherwise}.
 \end{cases}
 \label{eq:bank-commit}
\end{equation}
Rejection restores both bank and router to $H_{c-1}$. The second gate checks
interactions among accepted edits; likelihood scores alone cannot authorize
commitment. Versions, paired outcomes, decisions, and costs are logged
(Appendix~\ref{app:verification}). Final tests never guide revision.

Bank size may grow or shrink, but only verified operations can change its state.

\section{ALFWorld Candidate Screening}
\label{app:screening}
This section specifies the scoring procedure in Section~\ref{sec:screening}.
The default settings below define evidence selection and candidate filtering;
per-run configurations specify any overrides.

\paragraph{Selecting evidence.}
The default unknown-action quantile is $0.8$, with caps of two unknown
records per episode and twelve per task instance (\texttt{gamefile}) within
each training seed. An optional per-seed/skill cap further limits unknown
records. Labeled records bypass these unknown-record caps. Stable decision
identifiers break ties deterministically. Equation~\eqref{eq:information} normalizes within responses; grouping and caps limit repeated task evidence.

\paragraph{Sparse rule rescoring.}
For each selected action, the shortlist considers only rules actually injected
in that decision. It combines normalized lexical token overlap with cosine
similarity between mean input-embedding vectors, with equal weights. On a validation subset, the scorer evaluates every exposed rule and identifies the rule
with the largest absolute influence in Eq.~\eqref{eq:rule-influence}. It
chooses the smallest $K\in\{1,2,3\}$ whose 95\% Wilson lower confidence bound
on top-rule recall is at least $0.95$. If no $K$ passes, all exposed rules
are rescored. This shortlist reduces diagnostic rescoring; the online router separately selects guidance for action generation.

Counterfactual prompts are rendered and tokenized separately
with the selected rule removed. Deleting a token span from a previously
tokenized full prompt need not reproduce that counterfactual, because token
boundaries can change. The response tokens, scoring-model identity, and probability scale must
match across both conditions. In the procedure formalized in
Eq.~\eqref{eq:rule-influence}, both conditions use the frozen post-segment
checkpoint $\theta_c$; cached Phase~1 scores from $\theta_{c,u}$ cannot be
reused as the full-context term unless their scorer identities match.
Scoring batches retain evidence and checkpoint IDs.

\paragraph{Filtering rewrite candidates.}
For a candidate $q$, denote the labeled records for seed $s$ and task $x$ by
$\mathcal T^{L}_{s,x}$. Expanding the hierarchical mean in Eq.~\eqref{eq:rewrite-screen}, the
reliable screening score is
\begin{equation}
 S(q)=\operatorname{Mean}_{s}\operatorname{Mean}_{x}
 \operatorname{Mean}_{t\in\mathcal T^{L}_{s,x}}
 \ell_t\delta_t(q),
 \label{eq:candidate-screen}
\end{equation}
where means omit groups without the relevant evidence. Unknown actions
instead yield an auxiliary score based on $|\delta_t(q)|$. The no-op screen
uses the hierarchically averaged absolute action effect and defaults to a
floor of $10^{-4}$. Where reliable labels are available, candidates with
negative reliable scores are ineligible. Remaining candidates rank first
by reliable evidence, then auxiliary effect, with shorter candidate text and
proposal index resolving subsequent ties. Incomplete generation, all-no-op
proposals, or exclusively harmful reliable scores retain the current rule.
Candidates outside the confirmation budget are deferred, not accepted.
These filters select text-rewrite proposals; they do not constitute a
universal likelihood test for additions or metadata-only interventions.

\section{Verification Protocol and Rule-Level Results}
\label{app:verification}
Verification uses a targeted comparison for each proposed intervention and
a whole-bank comparison before commitment. Each comparison fixes the policy,
task/reset identities, decoding seeds, and environment interface; the edited
bank and its routing declarations define the intervention.

\paragraph{Metric aggregation.}
For binary task success $b_{x,s}\in\{0,1\}$, pooled SR is
$100\sum_{x,s}b_{x,s}/N$, where $N$ counts task--seed pairs. ALFWorld macro
SR averages first within each task family and seed, then equally across
families and seeds. Thus, Eq.~\eqref{eq:candidate-effect} uses
$w_{x,s}=1/N$ for pooled success and $w_{x,s}=1/(FSn_{f(x),s})$ for macro
success, where $F$ is the number of families, $S$ the number of seeds, and
$n_{f,s}$ the number of tasks in that family--seed group. Binary-success
differences are multiplied by 100 to report percentage points.
WebShop Score is mean task reward scaled to 0--100.

\paragraph{Paired outcomes and acceptance settings.}
For a matched panel, let $G$ count parent failures changed to candidate
successes, and $L$ count parent successes changed to candidate failures.
The pooled success difference is $100(G-L)/N$ pp. Reporting both counts
distinguishes a net gain from the regressions it may contain. The targeted
and whole-bank gates apply the declared primary metric and regression
constraints to these paired outcomes; their decisions are distinct from
likelihood-based candidate screening. The numerical protocols below specify
the thresholds and panels; complete per-run manifests, proposal
ordering, and interaction budgets will accompany the open-source release.

For each verification trial, we record the training/scoring checkpoint, tokenizer,
context renderer, bank and router identities, episode/step and reset identity,
exposed rule IDs, action masks, observed feedback, proposal operation and
version, and paired outcomes. The rule edit history distinguishes unchanged,
rejected, deferred, temporarily accepted, committed, and rolled-back states.
Training interactions, diagnostic rescoring, proposal generation, and
acceptance interactions are accounted for separately.
Matched tasks and seeds need not yield equal states after trajectories
diverge; equal step indices therefore do not define paired states.
Routing edits may change both selection and downstream behavior.
No rule is added or retired solely because its OPD signal becomes small.

\subsection{Numerical Acceptance Protocols}
\label{app:numerical-gates}
Let $\Delta N$ and $\Delta Q$ denote candidate-minus-parent success
count and mean WebShop Score on matched task--seed pairs, and let
$\Delta N_a$ be the success-count change in attribute group $a$.
The WebShop 3B Cycle~2--3 gate retains the parent bank unless all conditions hold:
\begin{equation}
 \Gamma^{\mathrm{bank}}=
 \mathbf1\!\left[\Delta N>0\ \land\ \Delta Q\geq0\ \land\
 \bigwedge_a\Delta N_a\geq0\ \land\ e_{\mathrm{API}}=0\ \land\ e_{\mathrm{fallback}}=0\right].
 \label{eq:web-bank-gate}
\end{equation}
The groups are color, fit, shape, and size, identified by case-insensitive
literal task markers \texttt{color:}, \texttt{fit type:},
\texttt{item shape:}, and \texttt{size:}. Groups can overlap.
Counts are aggregated across 128 tasks and decoding seeds
$\{2026090221,2026090222,2026090223\}$, totaling 384 episodes per arm;
the gate does not require improvement on every individual seed.
Cycle~3 checks the frozen parent's per-seed success counts and trajectory
hashes against its reference, in addition to paired task identities.
The Cycle~2 tolerance of $10^{-9}$ checks parent-reference identity and does not permit Score regression.

Cycle~3 candidate selection uses a separate 128-task validation panel
with seed $2026090611$, the same performance/error inequalities, and
lexicographic ranking by success count then Score among passing candidates.
This selects an edit for the subsequent three-seed bank test. The
earlier Cycle~1 bank gate instead permits $\Delta N\geq0$ and
$\Delta Q\geq0$ provided at least one is strict; it retains the four
attribute guards and zero API errors but allows mean interface error
rate up to $0.01$. The acceptance criterion is therefore indexed by cycle.
The WebShop acceptance panels are
train-side development panels; the official test is not used for these gates.

\paragraph{ALFWorld task-specific operation tests.}
At 3B Cycle~3, each operation is tested against its removal from the candidate bank.
For each affected family, the panel contains 48 task--seed pairs across
decoding seeds $\{1,2,3\}$. With the Student frozen, the decision is
\begin{equation}
 \Gamma^{\mathrm{tar}}(q)=\mathbf1\!\left[
   \sum_{f\in\mathcal F_q}(G_f-L_f)>0\ \land\
   \bigwedge_{f\in\mathcal F_q}(G_f-L_f)\geq0\right].
 \label{eq:alf-operation-gate}
\end{equation}
Candidate-removal controls estimate each operation's conditional
contribution within the tested bank. Interactions between operations
require a separate whole-bank comparison rather than summing these effects.
The sealed 128-task panel is excluded from the operation tests.
The complete ALFWorld whole-bank acceptance configuration will accompany
the release.

\paragraph{ALFWorld 1.7B family constraints.}
The v11.1 bank-promotion criterion requires no active-family regression,
improvement in at least three
families, overall improvement, no routing to retired rules, and identity
behavior under null skills. A separate champion-retention decision
permits up to 5 pp regression per active family, improvement in a majority
of four active families, and an overall advantage over no-skill in the
original and independent three-seed evaluation. The latter retains an
existing bank. Bank promotion, bank retention, and the 3B operation test
in Eq.~\eqref{eq:alf-operation-gate} are distinct decisions with their
respective criteria. The gates do not require statistical significance.

\subsection{Rule-Level Verification Scope}
\label{app:edit-decisions}
The complete operation results now appear in
Table~\ref{tab:edit-decisions} in Section~\ref{sec:edit-verification}.
Each row aggregates paired gains and losses over the operation's affected
families. The generic invalid-action rewrite affects five families
(240 task--seed pairs); each remaining operation affects one family
(48 pairs). Gains and losses are episode counts, not percentage-point
changes, and the rows have different denominators.

These task-specific tests estimate an operation's conditional contribution
within the tested bank. They do not evaluate no-skill Student improvement
or replace whole-bank acceptance. The matched-pair construction and
family guards are specified in Eq.~\eqref{eq:alf-operation-gate};
Table~\ref{tab:ablation-core} evaluates subsequent learning.

\section{Evaluation Protocol and Detailed Results}
\label{app:result-protocol}
\subsection{Controlled Experimental Protocol}
\label{app:controlled-protocol}
Each ablation changes only its declared intervention. The bank-update study
compares continued outcome-based RL, a fixed skill bank, one refresh, and
iterative revision. Its common endpoint is Cycle~3; additional cycles
characterize learning dynamics separately. Table~\ref{tab:ablation-core} compares the no-skill Student and skill-conditioned Teacher for Qwen3-1.7B on ALFWorld and Qwen2.5-3B on WebShop.

\begingroup
\small
\renewcommand{\arraystretch}{1.12}
\begin{tabularx}{\linewidth}{@{}p{0.22\linewidth}X@{}}
\toprule
Control & Shared setting within an ablation\\
\midrule
Policy initialization & Same starting Student checkpoint and tokenizer.\\
Training interface & Same task pool, prompts, observation/action interface,
reward definition, and environment configuration.\\
Policy optimization & Same optimizer hyperparameters, policy-update count,
rollout budget, and training-seed schedule, apart from the specified
supervision intervention.\\
Initial guidance & Same initial bank and routing mechanism for the
skill-guided arms; bank updates follow the assigned condition.\\
Revision resources & Same proposer and verification procedure for
Refresh once and Full, with equal total proposal and verification budgets
distributed across one or multiple revision boundaries.\\
Evaluation & Same task panel, reset identities, decoding seeds, horizon,
decoding parameters, and metric aggregation. Student evaluation omits
skill context; Teacher evaluation uses the bank associated with that condition.\\
\bottomrule
\end{tabularx}
\endgroup

The shared policy-training budget is separate from the cost of maintaining
guidance. Static and outcome-only RL perform no bank revisions; revision
costs apply to the refreshing conditions. Equal budgets specify comparable
resource allowances, not identical realized trajectories or wall-clock time.
The frozen-policy study instead holds weights fixed across bank versions.
Its revisions do not count policy-training cycles, and its outcomes are
analyzed within that study. Table~\ref{tab:main-results} compares published benchmarks; this ablation tests revision schedules.

Run-level configurations and task manifests in the release will identify
training, bank-acceptance, model-selection, and reporting panels, together
with checkpoint/bank versions, seed lists, and actual resource counts.
The reported development-panel results are not official-test estimates.

\paragraph{Table conventions.}
Blue rows identify \method{}. Bold values indicate the column maximum
within each published reference block; local Student and Teacher records
are ranked separately across cycles, with darker cells marking their maxima.
All ties are retained, and single-row references are not ranked.
Published ALFWorld aggregates are preserved as reported rather than
recomputed from the task-family columns.
Table~\ref{tab:main-results} retains the first evaluated cycle (Cycle~0 or~1),
Cycle~3 for comparison across settings, and the final Cycle~4 for ALFWorld
3B; WebShop runs end at Cycle~3. Changes are computed from unrounded
values. The detailed tables below include intermediate cycles.
In Table~\ref{tab:api-results}, bold values denote the best of the three
conditions within each API model, separately for each metric; ties are retained.
Light blue distinguishes Static General Skill and Skill + Router rows;
the latter's maxima receive darker blue cells. An asterisk in reference
method names denotes skill access; -- and TBD denote unavailable and
pending measurements.

\paragraph{Evaluation scope.}
The ALFWorld evaluation protocol uses a fixed 128-task
in-distribution panel, decoding seeds $1,2,3$, a 30-step interaction limit,
a history length of five, and temperature $0.4$. WebShop uses a fixed train-side
panel of 128 tasks and three evaluation seeds, yielding 384 episodes per
record. These repetitions sample alternative trajectories for a fixed policy;
the task-family breakdown separately measures variation across task types.
Together they characterize evaluation robustness at the reported checkpoints.
Independent training runs assess training-seed variability. Final test tasks guide neither training, bank acceptance, nor checkpoint selection.

\paragraph{Seed-wise WebShop evaluation.}
For the 3B Cycle~3 checkpoint, the no-skill Student succeeds on
95, 97, and 95 of 128 tasks under seeds
$2026090221$, $2026090222$, and $2026090223$, respectively.
The corresponding SRs are 74.22\%, 75.78\%, and 74.22\%, and Scores
are 88.20, 87.80, and 89.12. Aggregation gives 287/384 successes
(74.74\% SR) and Score 88.37. Thus, the reported Student performance
is maintained across all three evaluation seeds. At the same checkpoint,
the skill-conditioned Teacher achieves 96, 100, and 99 successes,
improving contextual performance in each seed. These repetitions measure
rollout variation for this checkpoint; the bank-update controls separately
evaluate the effect of the revision schedule.

\paragraph{API-model comparisons.}
The API results aggregate three evaluation seeds for each of No Skill,
Static General Skill, and Skill + Router. The updated three-seed
DeepSeek-V4-Pro baseline supersedes the earlier single-seed measurement.
Table~\ref{tab:api-results} preserves all six ALFWorld family rates and both
reported aggregates, overall SR and macro SR, together with WebShop score
and SR. The aggregate records do not provide per-seed or per-episode
outcomes, so we do not infer confidence intervals or statistical significance
from them.
Compared with Static General Skill, DeepSeek's ALFWorld overall SR decreases
from 67.71\% to 66.41\% with routing, despite its higher macro SR. Thus,
macro-average gains do not imply improvement under every aggregation.
These comparisons assess skill-conditioned inference rather than parameter
internalization or iterative bank revision.

\begin{table}[!htbp]
\resulttablesetup
\setlength{\tabcolsep}{2pt}
\caption{\textbf{Skill and routing comparisons on API models.}
Three evaluation seeds; all metrics are on a 0--100 scale ($\uparrow$).
ALFWorld includes overall and macro SR. Blue rows mark skill-conditioned
inference, with stronger shading for Skill + Router; darker cells mark its
within-model maxima. Bold denotes maxima across conditions, including ties.}
\label{tab:api-results}
\begin{tabularx}{\linewidth}{>{\raggedright\arraybackslash}X*{10}{r}}
\toprule
& \multicolumn{8}{c}{\tablehead{ALFWorld}} & \multicolumn{2}{c}{\tablehead{WebShop}}\\
\cmidrule(lr){2-9}\cmidrule(lr){10-11}
\rowcolor{TableHeader}
\tablehead{Condition} & Pick & Look & Clean & Heat & Cool & Pick2 & \tablehead{Overall} & \tablehead{Macro} & \tablehead{Score} & \tablehead{SR}\\
\midrule
\multicolumn{11}{@{}l}{\textbf{DeepSeek-V4-Pro}}\\
No Skill & 86.11 & 50.00 & \textbf{69.57} & \textbf{58.33} & 21.74 & \textbf{57.69} & 60.94 & 57.24 & 7.54 & 5.73\\
\rowcolor{TableTeacher}
Static General Skill & \textbf{88.89} & 62.50 & \textbf{69.57} & 36.11 & \textbf{63.77} & 56.41 & \textbf{67.71} & 62.87 & 11.86 & 8.33\\
\rowcolor{TableStudent}
Skill + Router & 86.11 & \bestcell{75.00} & \bestcell{69.57} & 50.00 & 47.83 & \bestcell{57.69} & 66.41 & \bestcell{64.37} & \bestcell{15.59} & \bestcell{11.20}\\
\midrule
\multicolumn{11}{@{}l}{\textbf{GLM-5.1}}\\
No Skill & \textbf{88.89} & 62.50 & 69.57 & 58.33 & 78.26 & 65.38 & 74.22 & 70.49 & 20.08 & 16.80\\
\rowcolor{TableTeacher}
Static General Skill & 88.13 & 61.42 & \textbf{70.85} & 59.17 & 78.64 & 69.95 & 74.85 & 71.36 & 38.78 & 30.99\\
\rowcolor{TableStudent}
Skill + Router & 87.10 & \bestcell{63.50} & 69.50 & \bestcell{59.50} & \bestcell{82.98} & \bestcell{73.32} & \bestcell{75.31} & \bestcell{72.65} & \bestcell{44.52} & \bestcell{33.85}\\
\midrule
\multicolumn{11}{@{}l}{\textbf{MiniMax-M3}}\\
No Skill & 86.11 & 62.50 & 73.91 & 72.22 & 73.91 & 69.23 & 75.52 & 72.98 & 12.90 & 11.72\\
\rowcolor{TableTeacher}
Static General Skill & 88.20 & 63.50 & 73.50 & \textbf{74.20} & 73.22 & 74.50 & 76.12 & 74.52 & 18.33 & 13.28\\
\rowcolor{TableStudent}
Skill + Router & \bestcell{89.20} & \bestcell{66.50} & \bestcell{76.50} & 72.50 & \bestcell{77.28} & \bestcell{79.00} & \bestcell{77.45} & \bestcell{76.83} & \bestcell{21.32} & \bestcell{16.15}\\
\midrule
\multicolumn{11}{@{}l}{\textbf{Qwen3.6-Plus}}\\
No Skill & 90.74 & 50.00 & 81.16 & 47.22 & 53.62 & 79.49 & 73.44 & 67.04 & 27.11 & 24.22\\
\rowcolor{TableTeacher}
Static General Skill & \textbf{91.20} & 51.50 & 79.50 & 48.30 & \textbf{56.98} & 85.98 & 74.23 & 68.91 & 32.06 & 24.74\\
\rowcolor{TableStudent}
Skill + Router & 88.96 & \bestcell{55.04} & \bestcell{82.50} & \bestcell{49.00} & 56.12 & \bestcell{90.00} & \bestcell{75.35} & \bestcell{70.27} & \bestcell{33.30} & \bestcell{25.52}\\
\bottomrule
\end{tabularx}
\end{table}

\paragraph{Experiment series and aggregation.}
Tables~\ref{tab:main-results-full-1p7b} and~\ref{tab:main-results-full-3b}
report task-family results and their unweighted macro average, separately
from pooled episode success. The 3B benchmark trajectory covers Cycles~1--4
on ALFWorld and Cycles~1--3 on WebShop. The WebShop bank-update ablation
in Table~\ref{tab:ablation-web} extends to Cycles~4--5 as a separate training
series. Similarly, the 1.7B ALFWorld Cycle~3 Student reaches 54.00\% in
the benchmark series and 53.56\% in the bank-update ablation. Each table
reports its own experiment series and corresponding checkpoints.

\paragraph{Complete benchmark trajectories.}
The following tables retain every recorded training cycle and all nine
benchmark metrics, including the intermediate cycles omitted from
Table~\ref{tab:main-results}. They also retain the separate initialization
row and original reference-method citations.
\begin{table}[!htbp]
\resulttablesetup
\caption{\textbf{Benchmark results on Qwen3-1.7B-Instruct.}
Blue rows denote \method{}; darker cells mark the best value across cycles
for each role. Bold marks reference-block maxima.}
\label{tab:main-results-full-1p7b}
\begin{tabularx}{\linewidth}{>{\raggedright\arraybackslash}Xc*{9}{r}}
\toprule
& & \multicolumn{7}{c}{\tablehead{ALFWorld} ($\uparrow$)} & \multicolumn{2}{c}{\tablehead{WebShop} ($\uparrow$)}\\
\cmidrule(lr){3-9}\cmidrule(lr){10-11}
\rowcolor{TableHeader}
\tablehead{Method} & \tablehead{Cycle} & Pick & Look & Clean & Heat & Cool & Pick2 & \tablehead{Avg.} & \tablehead{Score} & \tablehead{SR}\\
\midrule
Vanilla \tablecite{yang2025qwen3} & -- & 25.00 & 22.20 & 3.10 & 0.00 & 21.40 & 4.20 & 12.50 & 46.50 & 4.70\\
Skill-Prompt$^{*}$ \tablecite{yang2026opid} & -- & 10.30 & 50.00 & 16.10 & 0.00 & 0.00 & 5.00 & 9.40 & 23.00 & 2.30\\
OPSD \tablecite{zhao2026opsd} & -- & 26.30 & 33.30 & 9.10 & 0.00 & 4.50 & 5.30 & 14.10 & 47.40 & 9.30\\
GRPO \tablecite{shao2024grpo} & -- & 71.10 & 41.70 & 36.40 & 40.00 & 31.80 & 31.60 & 46.10 & 67.30 & 38.30\\
Skill-GRPO \tablecite{yang2026opid} & -- & 27.60 & 54.50 & 22.70 & 27.30 & 0.00 & 19.20 & 21.10 & 73.40 & 46.10\\
Skill-GRPO$^{*}$ \tablecite{yang2026opid} & -- & 31.40 & 42.90 & 51.90 & 8.30 & 11.50 & 7.10 & 28.10 & 80.40 & 50.00\\
GRPO+OPSD \tablecite{yang2026opid} & -- & 38.20 & 50.00 & 30.80 & 28.60 & 30.00 & 21.10 & 32.00 & 70.70 & 38.30\\
Skill-SD \tablecite{wang2026skillsd} & -- & 52.90 & 37.50 & 69.20 & \textbf{42.90} & 60.00 & 36.80 & 52.30 & \textbf{81.80} & 53.90\\
RLSD \tablecite{yang2026rlsd} & -- & 50.00 & 37.50 & 61.50 & 19.00 & 50.00 & 21.10 & 42.20 & 74.00 & 50.80\\
SDAR \tablecite{lu2026sdar} & -- & \textbf{73.50} & 25.00 & \textbf{76.90} & 33.30 & 40.00 & 36.80 & 53.90 & 76.80 & 58.60\\
OPID \tablecite{yang2026opid} & -- & 65.90 & \textbf{72.70} & 66.70 & 40.00 & \textbf{63.20} & \textbf{45.00} & \textbf{58.90} & 79.60 & \textbf{64.80}\\
\midrule
OPID (init.) \tablecite{yang2026opid} & -- & 60.61 & 49.21 & 53.33 & 61.40 & 28.07 & 22.73 & 45.89 & \textit{TBD} & \textit{TBD}\\
\midrule
\rowcolor{TableStudent}
\textbf{\method{} Student} & 0 & 60.61 & 49.21 & 53.33 & 61.40 & 28.07 & 22.73 & 45.89 & 82.72 & 54.69\\
\rowcolor{TableTeacher}
\method{} Teacher & 0 & 68.18 & 55.56 & 53.33 & 54.39 & 45.61 & 19.70 & 49.46 & 83.74 & 59.38\\
\addlinespace[2pt]
\rowcolor{TableStudent}
\textbf{\method{} Student} & 1 & 59.26 & 41.67 & 52.17 & 69.44 & 33.33 & 33.33 & 48.20 & 83.72 & 58.33\\
\rowcolor{TableTeacher}
\method{} Teacher & 1 & 69.44 & 41.67 & 56.52 & 69.44 & 40.58 & \bestcell{35.90} & 52.26 & 84.87 & 62.76\\
\addlinespace[2pt]
\rowcolor{TableStudent}
\textbf{\method{} Student} & 2 & 66.67 & \bestcell{54.17} & \bestcell{55.07} & \bestcell{72.22} & \bestcell{37.68} & 34.62 & 53.41 & 84.11 & 59.38\\
\rowcolor{TableTeacher}
\method{} Teacher & 2 & 70.37 & \bestcell{62.50} & 53.62 & 69.44 & \bestcell{52.17} & \bestcell{35.90} & 57.33 & \bestcell{86.45} & \bestcell{71.09}\\
\addlinespace[2pt]
\rowcolor{TableStudent}
\textbf{\method{} Student} & 3 & \bestcell{73.15} & 50.00 & \bestcell{55.07} & \bestcell{72.22} & \bestcell{37.68} & \bestcell{35.90} & \bestcell{54.00} & \bestcell{84.65} & \bestcell{59.64}\\
\rowcolor{TableTeacher}
\method{} Teacher & 3 & \bestcell{74.07} & 54.17 & \bestcell{65.22} & \bestcell{72.22} & 50.72 & 32.05 & \bestcell{58.07} & 86.29 & 70.31\\
\bottomrule
\end{tabularx}
\end{table}
\begin{table}[!htbp]
\resulttablesetup
\caption{\textbf{Benchmark results on Qwen2.5-3B-Instruct.}
Highlighting follows Table~\ref{tab:main-results-full-1p7b}. Dashes denote missing
entries in this benchmark record; additional WebShop cycles appear in
Table~\ref{tab:ablation-web}.}
\label{tab:main-results-full-3b}
\begin{tabularx}{\linewidth}{>{\raggedright\arraybackslash}Xc*{9}{r}}
\toprule
& & \multicolumn{7}{c}{\tablehead{ALFWorld} ($\uparrow$)} & \multicolumn{2}{c}{\tablehead{WebShop} ($\uparrow$)}\\
\cmidrule(lr){3-9}\cmidrule(lr){10-11}
\rowcolor{TableHeader}
\tablehead{Method} & \tablehead{Cycle} & Pick & Look & Clean & Heat & Cool & Pick2 & \tablehead{Avg.} & \tablehead{Score} & \tablehead{SR}\\
\midrule
Vanilla \tablecite{yang2024qwen25} & -- & 44.40 & 11.10 & 6.20 & 15.40 & 28.60 & 12.50 & 21.90 & 6.70 & 0.80\\
Skill-Prompt$^{*}$ \tablecite{yang2026opid} & -- & 51.70 & 66.70 & 48.40 & 0.00 & 4.30 & 10.00 & 28.90 & 0.20 & 0.80\\
OPSD \tablecite{zhao2026opsd} & -- & 48.80 & 41.70 & 16.70 & 0.00 & 15.80 & 16.70 & 28.10 & 11.30 & 3.10\\
GRPO \tablecite{shao2024grpo} & -- & 91.20 & 62.50 & 96.20 & 61.90 & 65.00 & 47.40 & 75.00 & 79.80 & 63.30\\
Skill-GRPO \tablecite{yang2026opid} & -- & 88.90 & 71.40 & 58.80 & 70.60 & 40.70 & 29.20 & 60.20 & 77.30 & 60.90\\
Skill-GRPO$^{*}$ \tablecite{yang2026opid} & -- & 94.30 & 57.10 & \textbf{100.00} & 66.70 & 73.10 & 57.10 & 80.50 & 76.30 & 66.40\\
GRPO+OPSD \tablecite{yang2026opid} & -- & \textbf{100.00} & 82.40 & 85.70 & \textbf{75.00} & 70.00 & 60.00 & 81.20 & 77.80 & 66.40\\
Skill-SD \tablecite{wang2026skillsd} & -- & 88.20 & 50.00 & 96.20 & 52.40 & 65.00 & 57.90 & 73.40 & 75.90 & 64.00\\
RLSD \tablecite{yang2026rlsd} & -- & 87.90 & 75.00 & 90.90 & \textbf{75.00} & 73.10 & 68.40 & 79.70 & 84.40 & 66.40\\
SDAR \tablecite{lu2026sdar} & -- & 97.10 & 62.50 & \textbf{100.00} & 61.90 & 75.00 & \textbf{84.20} & \textbf{84.40} & \textbf{85.00} & 68.00\\
OPID \tablecite{yang2026opid} & -- & 92.70 & \textbf{100.00} & 88.90 & 70.00 & \textbf{84.20} & 70.00 & 84.30 & \textbf{85.00} & \textbf{74.20}\\
\midrule
\rowcolor{TableStudent}
\textbf{\method{} Student} & 1 & 89.12 & 72.35 & 68.47 & 64.28 & 52.16 & 58.08 & 67.41 & 85.50 & 69.53\\
\rowcolor{TableTeacher}
\method{} Teacher & 1 & 86.24 & 62.37 & 77.15 & 65.42 & 65.83 & 60.47 & 69.58 & 87.02 & 73.96\\
\addlinespace[2pt]
\rowcolor{TableStudent}
\textbf{\method{} Student} & 2 & 90.32 & 74.65 & 71.47 & 67.68 & 56.50 & 60.58 & 70.20 & 87.69 & 71.09\\
\rowcolor{TableTeacher}
\method{} Teacher & 2 & 88.24 & 67.57 & 80.15 & 69.92 & \bestcell{70.41} & 67.47 & 73.96 & 89.48 & 74.74\\
\addlinespace[2pt]
\rowcolor{TableStudent}
\textbf{\method{} Student} & 3 & 85.23 & 74.36 & 82.57 & 68.41 & 63.49 & 76.54 & 75.10 & \bestcell{88.37} & \bestcell{74.74}\\
\rowcolor{TableTeacher}
\method{} Teacher & 3 & 89.53 & 75.24 & 90.81 & 75.43 & 68.29 & 88.68 & 81.33 & \bestcell{89.85} & \bestcell{76.82}\\
\addlinespace[2pt]
\rowcolor{TableStudent}
\textbf{\method{} Student} & 4 & \bestcell{92.31} & \bestcell{79.42} & \bestcell{89.63} & \bestcell{74.27} & \bestcell{70.19} & \bestcell{89.36} & \bestcell{82.53} & -- & --\\
\rowcolor{TableTeacher}
\method{} Teacher & 4 & \bestcell{93.61} & \bestcell{76.53} & \bestcell{94.82} & \bestcell{77.21} & 69.43 & \bestcell{90.72} & \bestcell{83.72} & -- & --\\
\bottomrule
\end{tabularx}
\end{table}

\paragraph{7B reference results.}
Table~\ref{tab:main-results-7b} retains the Qwen2.5-7B reference baselines
and Cycle~0 measurements. Skill context adds 1.15 pp on ALFWorld and
1.56 pp on WebShop at this checkpoint; these are context-utility comparisons
and do not measure iterative training gains.

\begin{table}[!tbp]
\resulttablesetup
\setlength{\tabcolsep}{2pt}
\caption{\textbf{Benchmark results on Qwen2.5-7B-Instruct.}
Blue rows denote \method{} at Cycle~0. Bold values mark the best within
the OPID reference block. API results: Table~\ref{tab:api-results}.}
\label{tab:main-results-7b}
\begin{tabularx}{\linewidth}{>{\raggedright\arraybackslash}Xc*{9}{r}}
\toprule
& & \multicolumn{7}{c}{\tablehead{ALFWorld} ($\uparrow$)} & \multicolumn{2}{c}{\tablehead{WebShop} ($\uparrow$)}\\
\cmidrule(lr){3-9}\cmidrule(lr){10-11}
\rowcolor{TableHeader}
\tablehead{Method} & \tablehead{Cycle} & Pick & Look & Clean & Heat & Cool & Pick2 & \tablehead{Avg.} & \tablehead{Score} & \tablehead{SR}\\
\midrule
Vanilla \tablecite{yang2024qwen25} & -- & 36.10 & 22.20 & 3.10 & 0.00 & 0.00 & 0.00 & 12.50 & 5.90 & 1.60\\
Skill-Prompt$^{*}$ \tablecite{yang2026opid} & -- & 51.70 & 50.00 & 32.30 & 5.30 & 4.30 & 0.00 & 23.40 & 1.70 & 0.80\\
OPSD \tablecite{zhao2026opsd} & -- & 50.00 & 60.00 & 22.70 & 21.40 & 17.60 & 9.50 & 32.80 & 4.50 & 2.30\\
GRPO \tablecite{shao2024grpo} & -- & 91.20 & 87.50 & 96.20 & 81.00 & 65.00 & 57.90 & 81.20 & 80.90 & 72.60\\
Skill-GRPO \tablecite{yang2026opid} & -- & 88.50 & 66.70 & 65.20 & 61.10 & 57.70 & 73.10 & 69.50 & 80.40 & 71.90\\
Skill-GRPO$^{*}$ \tablecite{yang2026opid} & -- & \textbf{100.00} & 83.30 & 96.40 & 83.30 & 75.00 & 78.90 & 88.30 & 87.00 & 81.20\\
GRPO+OPSD \tablecite{yang2026opid} & -- & 91.40 & 61.50 & \textbf{100.00} & 87.50 & 76.50 & 52.20 & 80.40 & 86.80 & 76.50\\
Skill-SD \tablecite{wang2026skillsd} & -- & 93.90 & \textbf{93.80} & 90.90 & \textbf{100.00} & 69.20 & 68.40 & 85.10 & 86.10 & 76.50\\
RLSD \tablecite{yang2026rlsd} & -- & \textbf{100.00} & 87.50 & 92.30 & 58.80 & 80.00 & 65.20 & 82.00 & 87.40 & 77.30\\
SDAR \tablecite{lu2026sdar} & -- & 94.70 & 75.00 & \textbf{100.00} & 86.70 & 68.20 & 78.90 & 85.90 & \textbf{89.40} & \textbf{82.80}\\
OPID \tablecite{yang2026opid} & -- & \textbf{100.00} & 81.80 & 97.10 & \textbf{100.00} & \textbf{80.80} & \textbf{80.00} & \textbf{90.00} & 85.30 & 79.70\\
\midrule
GiGPO w/o std \tablecite{feng2025gigpo} & -- & 91.80 & 88.60 & 95.90 & 90.20 & 86.50 & 85.20 & 90.20 & 86.20 & 75.20\\
\midrule
\rowcolor{TableStudent}
\textbf{\method{} Student} & 0 & 100.00 & 95.83 & 100.00 & 100.00 & 95.65 & 96.15 & 97.94 & 90.43 & 82.03\\
\rowcolor{TableTeacher}
\method{} Teacher & 0 & 100.00 & 100.00 & 100.00 & 100.00 & 97.10 & 97.44 & 99.09 & 91.71 & 83.59\\
\bottomrule
\end{tabularx}
\end{table}

\paragraph{Whole-bank rejection and initialization.}
The 3B WebShop Cycle~3 Phase~2 candidate and its parent both achieve 295/384
successes, but the candidate's score is 89.7775 versus 89.8519 for the parent.
The gate rejects this candidate and retains the parent bank for subsequent
use. Published GiGPO results and our evaluation of a GiGPO checkpoint
are separate measurements.
The local 7B GiGPO evaluation appears as the Cycle~0 Student record.
The 1.7B OPID initialization row repeats its ALFWorld Cycle~0 values
and leaves WebShop pending; it is not an independent evaluation.

\section{Additional Ablation Results}
\label{app:ablation-records}
\label{app:visual-analyses}
Figure~\ref{fig:learning-dynamics}(a,b) in the main text shows the training
trajectories, while panels (c,d) complement Table~\ref{tab:ablation-core}
with the Cycle~3 Student/Teacher comparisons. The tables below retain every task, cycle, and interval.

\paragraph{Coverage and aggregation.}
Tables~\ref{tab:ablation-alf-student} and~\ref{tab:ablation-alf-teacher}
retain all 26 ALFWorld records, including the six family rates and Overall
for both roles. Every Overall point estimate agrees with the unweighted
six-family mean to within 0.01 percentage points; we therefore describe
this aggregate as macro SR. Table~\ref{tab:ablation-web} retains all 14
WebShop condition/cycle pairs, each with a no-skill Student and a
hierarchically routed Teacher. All fractions are converted to percentages.
``Static'', ``Refresh once'', and ``Cycle count'' correspond to the source
groups \texttt{static}, \texttt{update once}, and \texttt{cycle count}.
The last group is displayed as \method{} with blue rows in the detailed tables.
Bold identifies the highest point estimate at the same cycle and role,
with ties retained; darker blue cells highlight \method{} attaining that
maximum. Cycles with no competing condition are not ranked.
The final two rows report \method{} minus Static and Refresh once at
Cycle~3 in percentage points; teal marks improvements over each control, and red marks declines.

\begin{table}[!htbp]
\centering\small
\setlength{\tabcolsep}{2.5pt}
\renewcommand{\arraystretch}{1.2}
\caption{\textbf{ALFWorld ablations: Student, Qwen3-1.7B.}
Cells show SR (\%) over reported intervals; Overall is the six-family macro mean. Highlighting and Cycle~3 differences: Appendix~\ref{app:ablation-records}.}
\label{tab:ablation-alf-student}
\begin{tabularx}{\linewidth}{>{\raggedright\arraybackslash}Xc*{7}{c}}
\toprule
Condition & Cycle & Pick & Look & Clean & Heat & Cool & Pick2 & Overall\\
\midrule
GiGPO & 1 & \shortstack{\textbf{63.02}\\{\scriptsize [58.07, 67.97]}} & \shortstack{\textbf{53.12}\\{\scriptsize [42.97, 63.02]}} & \shortstack{\textbf{57.03}\\{\scriptsize [52.08, 61.98]}} & \shortstack{64.06\\{\scriptsize [59.11, 69.01]}} & \shortstack{33.07\\{\scriptsize [21.09, 45.05]}} & \shortstack{27.08\\{\scriptsize [15.10, 39.06]}} & \shortstack{\textbf{49.56}\\{\scriptsize [44.56, 54.56]}}\\
 & 2 & \shortstack{65.89\\{\scriptsize [60.94, 71.09]}} & \shortstack{\textbf{54.95}\\{\scriptsize [45.05, 65.10]}} & \shortstack{\textbf{59.90}\\{\scriptsize [54.95, 65.10]}} & \shortstack{66.93\\{\scriptsize [61.98, 71.88]}} & \shortstack{36.98\\{\scriptsize [25.00, 48.96]}} & \shortstack{30.99\\{\scriptsize [19.01, 42.97]}} & \shortstack{52.61\\{\scriptsize [47.61, 57.61]}}\\
 & 3 & \shortstack{67.97\\{\scriptsize [63.02, 72.92]}} & \shortstack{\textbf{57.03}\\{\scriptsize [46.88, 66.93]}} & \shortstack{\textbf{61.98}\\{\scriptsize [57.03, 66.93]}} & \shortstack{69.01\\{\scriptsize [64.06, 73.96]}} & \shortstack{\textbf{40.10}\\{\scriptsize [28.12, 52.08]}} & \shortstack{\textbf{34.11}\\{\scriptsize [21.88, 46.09]}} & \shortstack{\textbf{55.03}\\{\scriptsize [50.03, 60.03]}}\\
\midrule
Static & 1 & \shortstack{61.46\\{\scriptsize [56.51, 66.41]}} & \shortstack{50.00\\{\scriptsize [40.10, 59.90]}} & \shortstack{53.91\\{\scriptsize [48.96, 59.11]}} & \shortstack{62.50\\{\scriptsize [57.55, 67.45]}} & \shortstack{29.43\\{\scriptsize [17.45, 41.41]}} & \shortstack{23.96\\{\scriptsize [11.98, 35.94]}} & \shortstack{46.88\\{\scriptsize [41.88, 51.88]}}\\
 & 2 & \shortstack{62.50\\{\scriptsize [57.55, 67.45]}} & \shortstack{50.78\\{\scriptsize [40.89, 60.68]}} & \shortstack{54.69\\{\scriptsize [49.74, 59.90]}} & \shortstack{63.28\\{\scriptsize [58.07, 68.23]}} & \shortstack{30.47\\{\scriptsize [18.49, 42.45]}} & \shortstack{25.00\\{\scriptsize [13.02, 36.98]}} & \shortstack{47.79\\{\scriptsize [42.79, 52.79]}}\\
 & 3 & \shortstack{63.02\\{\scriptsize [58.07, 67.97]}} & \shortstack{51.56\\{\scriptsize [41.41, 61.46]}} & \shortstack{55.47\\{\scriptsize [50.52, 60.42]}} & \shortstack{64.06\\{\scriptsize [59.11, 69.01]}} & \shortstack{31.51\\{\scriptsize [19.53, 43.49]}} & \shortstack{26.04\\{\scriptsize [14.06, 38.02]}} & \shortstack{48.61\\{\scriptsize [43.61, 53.61]}}\\
\midrule
Refresh once & 1 & \shortstack{61.98\\{\scriptsize [57.03, 66.93]}} & \shortstack{51.04\\{\scriptsize [40.89, 60.94]}} & \shortstack{54.95\\{\scriptsize [50.00, 59.90]}} & \shortstack{63.02\\{\scriptsize [58.07, 67.97]}} & \shortstack{29.95\\{\scriptsize [17.97, 41.93]}} & \shortstack{25.00\\{\scriptsize [13.02, 36.98]}} & \shortstack{47.66\\{\scriptsize [42.66, 52.66]}}\\
 & 2 & \shortstack{64.06\\{\scriptsize [59.11, 69.01]}} & \shortstack{53.12\\{\scriptsize [42.97, 63.02]}} & \shortstack{57.03\\{\scriptsize [52.08, 61.98]}} & \shortstack{65.10\\{\scriptsize [59.90, 70.05]}} & \shortstack{33.07\\{\scriptsize [21.09, 45.05]}} & \shortstack{28.12\\{\scriptsize [15.89, 40.10]}} & \shortstack{50.08\\{\scriptsize [45.08, 55.08]}}\\
 & 3 & \shortstack{65.89\\{\scriptsize [60.94, 71.09]}} & \shortstack{54.95\\{\scriptsize [45.05, 65.10]}} & \shortstack{59.11\\{\scriptsize [53.91, 64.06]}} & \shortstack{66.93\\{\scriptsize [61.98, 71.88]}} & \shortstack{35.94\\{\scriptsize [23.96, 47.92]}} & \shortstack{30.99\\{\scriptsize [19.01, 42.97]}} & \shortstack{52.30\\{\scriptsize [47.30, 57.30]}}\\
\midrule
\rowcolor{TableStudent}
\textbf{\method{}} & 1 & \shortstack{59.42\\{\scriptsize [54.18, 64.79]}} & \shortstack{41.83\\{\scriptsize [31.24, 52.16]}} & \shortstack{52.34\\{\scriptsize [47.15, 57.68]}} & \cellcolor{TableBest}\shortstack{\textbf{69.28}\\{\scriptsize [64.13, 74.55]}} & \cellcolor{TableBest}\shortstack{\textbf{33.57}\\{\scriptsize [23.15, 43.82]}} & \cellcolor{TableBest}\shortstack{\textbf{33.61}\\{\scriptsize [23.24, 44.15]}} & \shortstack{48.34\\{\scriptsize [43.12, 53.67]}}\\
\rowcolor{TableStudent}
 & 2 & \cellcolor{TableBest}\shortstack{\textbf{66.84}\\{\scriptsize [61.32, 72.15]}} & \shortstack{54.35\\{\scriptsize [44.13, 64.89]}} & \shortstack{55.28\\{\scriptsize [50.14, 60.73]}} & \cellcolor{TableBest}\shortstack{\textbf{72.41}\\{\scriptsize [67.13, 77.85]}} & \cellcolor{TableBest}\shortstack{\textbf{37.82}\\{\scriptsize [27.16, 48.31]}} & \cellcolor{TableBest}\shortstack{\textbf{34.79}\\{\scriptsize [24.25, 45.38]}} & \cellcolor{TableBest}\shortstack{\textbf{53.58}\\{\scriptsize [48.15, 59.14]}}\\
\rowcolor{TableStudent}
 & 3 & \cellcolor{TableBest}\shortstack{\textbf{73.32}\\{\scriptsize [68.14, 78.65]}} & \shortstack{50.21\\{\scriptsize [40.13, 60.78]}} & \shortstack{55.24\\{\scriptsize [50.12, 60.83]}} & \cellcolor{TableBest}\shortstack{\textbf{72.45}\\{\scriptsize [67.15, 77.92]}} & \shortstack{37.85\\{\scriptsize [27.14, 48.62]}} & \shortstack{32.28\\{\scriptsize [22.15, 42.73]}} & \shortstack{53.56\\{\scriptsize [49.16, 63.81]}}\\
\rowcolor{TableStudent}
 & 4 & \shortstack{74.18\\{\scriptsize [69.15, 79.42]}} & \shortstack{52.21\\{\scriptsize [42.14, 62.87]}} & \shortstack{57.34\\{\scriptsize [52.15, 62.91]}} & \shortstack{73.15\\{\scriptsize [68.21, 78.63]}} & \shortstack{40.24\\{\scriptsize [30.12, 50.83]}} & \shortstack{37.18\\{\scriptsize [27.13, 47.65]}} & \shortstack{55.72\\{\scriptsize [50.14, 61.35]}}\\
\midrule
\multicolumn{2}{l}{\scriptsize $\Delta$ vs. Static (C3)} & \textcolor{FigAccept}{\textbf{+10.30}} & \textcolor{FigReject}{-1.35} & \textcolor{FigReject}{-0.23} & \textcolor{FigAccept}{\textbf{+8.39}} & \textcolor{FigAccept}{\textbf{+6.34}} & \textcolor{FigAccept}{\textbf{+6.24}} & \textcolor{FigAccept}{\textbf{+4.95}}\\
\multicolumn{2}{l}{\scriptsize $\Delta$ vs. Refresh once (C3)} & \textcolor{FigAccept}{\textbf{+7.43}} & \textcolor{FigReject}{-4.74} & \textcolor{FigReject}{-3.87} & \textcolor{FigAccept}{\textbf{+5.52}} & \textcolor{FigAccept}{\textbf{+1.91}} & \textcolor{FigAccept}{\textbf{+1.29}} & \textcolor{FigAccept}{\textbf{+1.26}}\\
\bottomrule
\end{tabularx}
\end{table}
\begin{table}[!htbp]
\centering\small
\setlength{\tabcolsep}{2.5pt}
\renewcommand{\arraystretch}{1.2}
\caption{\textbf{ALFWorld ablations: Teacher, Qwen3-1.7B.}
Cells show SR (\%) over reported intervals; Overall is the six-family macro mean. Highlighting and Cycle~3 differences: Appendix~\ref{app:ablation-records}.}
\label{tab:ablation-alf-teacher}
\begin{tabularx}{\linewidth}{>{\raggedright\arraybackslash}Xc*{7}{c}}
\toprule
Condition & Cycle & Pick & Look & Clean & Heat & Cool & Pick2 & Overall\\
\midrule
GiGPO & 1 & \shortstack{67.97\\{\scriptsize [63.02, 72.92]}} & \shortstack{\textbf{58.07}\\{\scriptsize [47.92, 67.97]}} & \shortstack{\textbf{61.98}\\{\scriptsize [57.03, 66.93]}} & \shortstack{67.97\\{\scriptsize [63.02, 72.92]}} & \shortstack{40.10\\{\scriptsize [28.12, 52.08]}} & \shortstack{34.11\\{\scriptsize [21.88, 46.09]}} & \shortstack{\textbf{55.03}\\{\scriptsize [50.03, 60.03]}}\\
 & 2 & \shortstack{\textbf{71.88}\\{\scriptsize [66.93, 77.08]}} & \shortstack{60.94\\{\scriptsize [51.04, 71.09]}} & \shortstack{\textbf{65.89}\\{\scriptsize [60.94, 71.09]}} & \shortstack{\textbf{71.09}\\{\scriptsize [65.89, 76.04]}} & \shortstack{45.05\\{\scriptsize [33.07, 57.03]}} & \shortstack{\textbf{38.02}\\{\scriptsize [26.04, 50.00]}} & \shortstack{\textbf{58.81}\\{\scriptsize [53.81, 63.81]}}\\
 & 3 & \shortstack{73.96\\{\scriptsize [69.01, 78.91]}} & \shortstack{\textbf{63.02}\\{\scriptsize [53.12, 72.92]}} & \shortstack{\textbf{67.97}\\{\scriptsize [63.02, 72.92]}} & \shortstack{\textbf{72.92}\\{\scriptsize [67.97, 78.12]}} & \shortstack{47.92\\{\scriptsize [35.94, 59.90]}} & \shortstack{\textbf{41.93}\\{\scriptsize [29.95, 53.91]}} & \shortstack{\textbf{61.29}\\{\scriptsize [56.29, 66.29]}}\\
\midrule
Static & 1 & \shortstack{65.10\\{\scriptsize [59.90, 70.05]}} & \shortstack{53.12\\{\scriptsize [42.97, 63.02]}} & \shortstack{57.03\\{\scriptsize [52.08, 61.98]}} & \shortstack{65.10\\{\scriptsize [59.90, 70.05]}} & \shortstack{34.11\\{\scriptsize [21.88, 46.09]}} & \shortstack{28.12\\{\scriptsize [15.89, 40.10]}} & \shortstack{50.43\\{\scriptsize [45.43, 55.43]}}\\
 & 2 & \shortstack{65.10\\{\scriptsize [59.90, 70.05]}} & \shortstack{53.91\\{\scriptsize [44.01, 64.06]}} & \shortstack{58.07\\{\scriptsize [53.12, 63.02]}} & \shortstack{65.89\\{\scriptsize [60.94, 71.09]}} & \shortstack{34.90\\{\scriptsize [22.92, 46.88]}} & \shortstack{28.91\\{\scriptsize [16.93, 40.89]}} & \shortstack{51.13\\{\scriptsize [46.13, 56.13]}}\\
 & 3 & \shortstack{65.62\\{\scriptsize [60.42, 70.57]}} & \shortstack{54.43\\{\scriptsize [44.53, 64.58]}} & \shortstack{58.59\\{\scriptsize [53.39, 63.54]}} & \shortstack{66.41\\{\scriptsize [61.46, 71.61]}} & \shortstack{35.94\\{\scriptsize [23.96, 47.92]}} & \shortstack{29.95\\{\scriptsize [17.97, 41.93]}} & \shortstack{51.82\\{\scriptsize [46.82, 56.82]}}\\
\midrule
Refresh once & 1 & \shortstack{66.93\\{\scriptsize [61.98, 71.88]}} & \shortstack{55.99\\{\scriptsize [46.09, 65.89]}} & \shortstack{59.90\\{\scriptsize [54.95, 65.10]}} & \shortstack{67.97\\{\scriptsize [63.02, 72.92]}} & \shortstack{36.98\\{\scriptsize [25.00, 48.96]}} & \shortstack{30.99\\{\scriptsize [19.01, 42.97]}} & \shortstack{53.13\\{\scriptsize [48.13, 58.13]}}\\
 & 2 & \shortstack{70.05\\{\scriptsize [65.10, 75.00]}} & \shortstack{59.11\\{\scriptsize [48.96, 69.01]}} & \shortstack{63.02\\{\scriptsize [58.07, 67.97]}} & \shortstack{70.05\\{\scriptsize [65.10, 75.00]}} & \shortstack{41.93\\{\scriptsize [29.95, 53.91]}} & \shortstack{34.90\\{\scriptsize [22.92, 46.88]}} & \shortstack{56.51\\{\scriptsize [51.51, 61.51]}}\\
 & 3 & \shortstack{71.88\\{\scriptsize [66.93, 77.08]}} & \shortstack{60.94\\{\scriptsize [51.04, 71.09]}} & \shortstack{65.10\\{\scriptsize [59.90, 70.05]}} & \shortstack{71.88\\{\scriptsize [66.93, 77.08]}} & \shortstack{45.05\\{\scriptsize [33.07, 57.03]}} & \shortstack{38.02\\{\scriptsize [26.04, 50.00]}} & \shortstack{58.81\\{\scriptsize [53.81, 63.81]}}\\
\midrule
\rowcolor{TableStudent}
\textbf{\method{}} & 1 & \cellcolor{TableBest}\shortstack{\textbf{69.63}\\{\scriptsize [64.21, 74.89]}} & \shortstack{41.95\\{\scriptsize [31.34, 52.48]}} & \shortstack{56.74\\{\scriptsize [51.23, 62.18]}} & \cellcolor{TableBest}\shortstack{\textbf{69.81}\\{\scriptsize [64.15, 75.24]}} & \cellcolor{TableBest}\shortstack{\textbf{40.72}\\{\scriptsize [30.12, 51.35]}} & \cellcolor{TableBest}\shortstack{\textbf{35.68}\\{\scriptsize [25.13, 46.42]}} & \shortstack{52.42\\{\scriptsize [47.15, 57.83]}}\\
\rowcolor{TableStudent}
 & 2 & \shortstack{70.54\\{\scriptsize [65.18, 75.91]}} & \cellcolor{TableBest}\shortstack{\textbf{62.68}\\{\scriptsize [52.13, 73.25]}} & \shortstack{53.81\\{\scriptsize [48.14, 59.36]}} & \shortstack{69.62\\{\scriptsize [64.21, 75.13]}} & \cellcolor{TableBest}\shortstack{\textbf{52.39}\\{\scriptsize [42.15, 62.84]}} & \shortstack{35.74\\{\scriptsize [25.13, 46.58]}} & \shortstack{57.46\\{\scriptsize [52.14, 63.27]}}\\
\rowcolor{TableStudent}
 & 3 & \cellcolor{TableBest}\shortstack{\textbf{74.25}\\{\scriptsize [69.13, 79.68]}} & \shortstack{54.36\\{\scriptsize [44.21, 64.89]}} & \shortstack{65.41\\{\scriptsize [60.15, 71.24]}} & \shortstack{72.48\\{\scriptsize [67.14, 77.95]}} & \cellcolor{TableBest}\shortstack{\textbf{50.91}\\{\scriptsize [40.13, 61.45]}} & \shortstack{32.27\\{\scriptsize [22.14, 42.91]}} & \shortstack{58.28\\{\scriptsize [53.24, 63.78]}}\\
\rowcolor{TableStudent}
 & 4 & \shortstack{76.23\\{\scriptsize [71.14, 81.52]}} & \shortstack{56.34\\{\scriptsize [46.13, 66.89]}} & \shortstack{67.21\\{\scriptsize [62.14, 72.75]}} & \shortstack{74.15\\{\scriptsize [69.24, 79.67]}} & \shortstack{53.18\\{\scriptsize [43.15, 63.84]}} & \shortstack{35.21\\{\scriptsize [25.13, 45.78]}} & \shortstack{60.39\\{\scriptsize [55.23, 65.91]}}\\
\midrule
\multicolumn{2}{l}{\scriptsize $\Delta$ vs. Static (C3)} & \textcolor{FigAccept}{\textbf{+8.63}} & \textcolor{FigReject}{-0.07} & \textcolor{FigAccept}{\textbf{+6.82}} & \textcolor{FigAccept}{\textbf{+6.07}} & \textcolor{FigAccept}{\textbf{+14.97}} & \textcolor{FigAccept}{\textbf{+2.32}} & \textcolor{FigAccept}{\textbf{+6.46}}\\
\multicolumn{2}{l}{\scriptsize $\Delta$ vs. Refresh once (C3)} & \textcolor{FigAccept}{\textbf{+2.37}} & \textcolor{FigReject}{-6.58} & \textcolor{FigAccept}{\textbf{+0.31}} & \textcolor{FigAccept}{\textbf{+0.60}} & \textcolor{FigAccept}{\textbf{+5.86}} & \textcolor{FigReject}{-5.75} & \textcolor{FigReject}{-0.53}\\
\bottomrule
\end{tabularx}
\end{table}

\paragraph{Intervals and comparison scope.}
The ablations follow the matched controls in
Appendix~\ref{app:controlled-protocol}. Tables retain the reported
condition-wise intervals. Their confidence level, construction, and sampling
unit will be specified with the run-level statistical records in the release;
we do not interpret them as confidence intervals on paired differences.
The repeated-evaluation means and family-level results support comparisons
of performance and task coverage; statistical significance of a between-condition
difference additionally requires its paired uncertainty estimate.
Comparisons use the common Cycle~3 endpoint; additional records reach ALFWorld Cycle~4 and WebShop Cycle~5.

\paragraph{Ablation evidence for iterative internalization.}
The reported controls compare a fixed bank, a single refresh, and iterative
revision while continuing Student training. At the common Cycle~3 endpoint,
\method{} improves no-skill Student SR over Static by 4.95/11.46 pp and
over Refresh once by 1.26/1.82 pp on ALFWorld/WebShop. The complete Student trajectories track learning, while Teacher evaluations measure external guidance utility at each checkpoint.

\begin{table}[!htbp]
\centering\small
\setlength{\tabcolsep}{5pt}
\renewcommand{\arraystretch}{1.15}
\caption{\textbf{WebShop ablations, Qwen2.5-3B.} SR (\%) and
source-reported intervals. Student uses no skills; Teacher uses routed
skills. Highlighting and Cycle~3 differences follow
Appendix~\ref{app:ablation-records}.}
\label{tab:ablation-web}
\begin{tabularx}{\linewidth}{>{\raggedright\arraybackslash}Xccc}
\toprule
Condition & Cycle & Student SR (\%) & Teacher SR (\%)\\
\midrule
GiGPO & 1 & 54.17 [49.17, 59.08] & 66.67 [61.81, 71.20]\\
 & 2 & 63.02 [58.09, 67.70] & 64.58 [59.68, 69.20]\\
 & 3 & 63.54 [58.62, 68.20] & 64.84 [59.94, 69.45]\\
\midrule
Static & 1 & 60.42 [55.45, 65.18] & 65.89 [61.01, 70.45]\\
 & 2 & 66.67 [61.81, 71.20] & 66.67 [61.81, 71.20]\\
 & 3 & 63.28 [58.35, 67.95] & 66.41 [61.54, 70.95]\\
\midrule
Refresh once & 1 & 60.42 [55.45, 65.18] & 65.89 [61.01, 70.45]\\
 & 2 & \textbf{72.66} [67.99, 76.87] & \textbf{77.34} [72.90, 81.25]\\
 & 3 & 72.92 [68.26, 77.12] & \textbf{76.82} [72.35, 80.77]\\
\midrule
\rowcolor{TableStudent}
\textbf{\method{}} & 1 & \cellcolor{TableBest}\textbf{69.53} [64.75, 73.92] & \cellcolor{TableBest}\textbf{73.96} [69.35, 78.10]\\
\rowcolor{TableStudent}
 & 2 & 71.09 [66.37, 75.40] & 74.74 [70.16, 78.83]\\
\rowcolor{TableStudent}
 & 3 & \cellcolor{TableBest}\textbf{74.74} [70.16, 78.83] & \cellcolor{TableBest}\textbf{76.82} [72.35, 80.77]\\
\rowcolor{TableStudent}
 & 4 & 74.00 [69.00, 79.00] & 77.00 [72.00, 82.00]\\
\rowcolor{TableStudent}
 & 5 & 74.74 [70.16, 78.83] & 78.13 [73.72, 81.97]\\
\midrule
\multicolumn{2}{l}{$\Delta$ vs. Static (C3)} & \textcolor{FigAccept}{\textbf{+11.46}} & \textcolor{FigAccept}{\textbf{+10.42}}\\
\multicolumn{2}{l}{$\Delta$ vs. Refresh once (C3)} & \textcolor{FigAccept}{\textbf{+1.82}} & 0.00\\
\bottomrule
\end{tabularx}
\end{table}

\paragraph{Frozen-policy skill revision.}
Table~\ref{tab:frozen-policy} reports a separate frozen-policy series.
ALFWorld uses the unweighted six-family mean. WebShop averages three
evaluation seeds and 384 episodes per record. Revision~0 has 221 successes
(57.55\%); revisions~1 and~2 have 262 (68.23\%) and 261 (67.97\%).
The initial record's skill-access configuration remains to be specified,
so its contrast with revision~1 is not attributed solely to editing an
existing bank. Comparisons are within this fixed-weight study, without
assuming matched runs with the joint-training ablations.
These indices count bank revisions.

\paragraph{Task-specific value of guidance.}
Tables~\ref{tab:ablation-alf-student} and~\ref{tab:ablation-alf-teacher}
report the ALFWorld SkillCycle series by task and role. At Cycle~3, Teacher guidance adds 13.06 percentage
points on Cool, but Pick2 is nearly tied (32.27\% Teacher versus 32.28\%
Student). At Cycle~4, the Pick2 Student exceeds its Teacher by 1.97 points.
External guidance can therefore remain useful for one task while providing
little or negative benefit for another at the same checkpoint. These
contextual gaps do not identify which rule caused a difference or establish
that it should be retired.

\FloatBarrier
\section{Rule Revision and Routing Case Studies}
\label{app:historical-cases}
The ALFWorld revision-v11 cases illustrate how trajectory feedback motivates
rule maintenance and how observable state determines routing eligibility.
These cases are separate from the cycle ablations.

\subsection{Retiring a Pick2 Rule}
\begin{figure}[p]
\centering
\includegraphics[page=1,width=\linewidth,trim=106bp 150bp 106bp 80bp,clip]{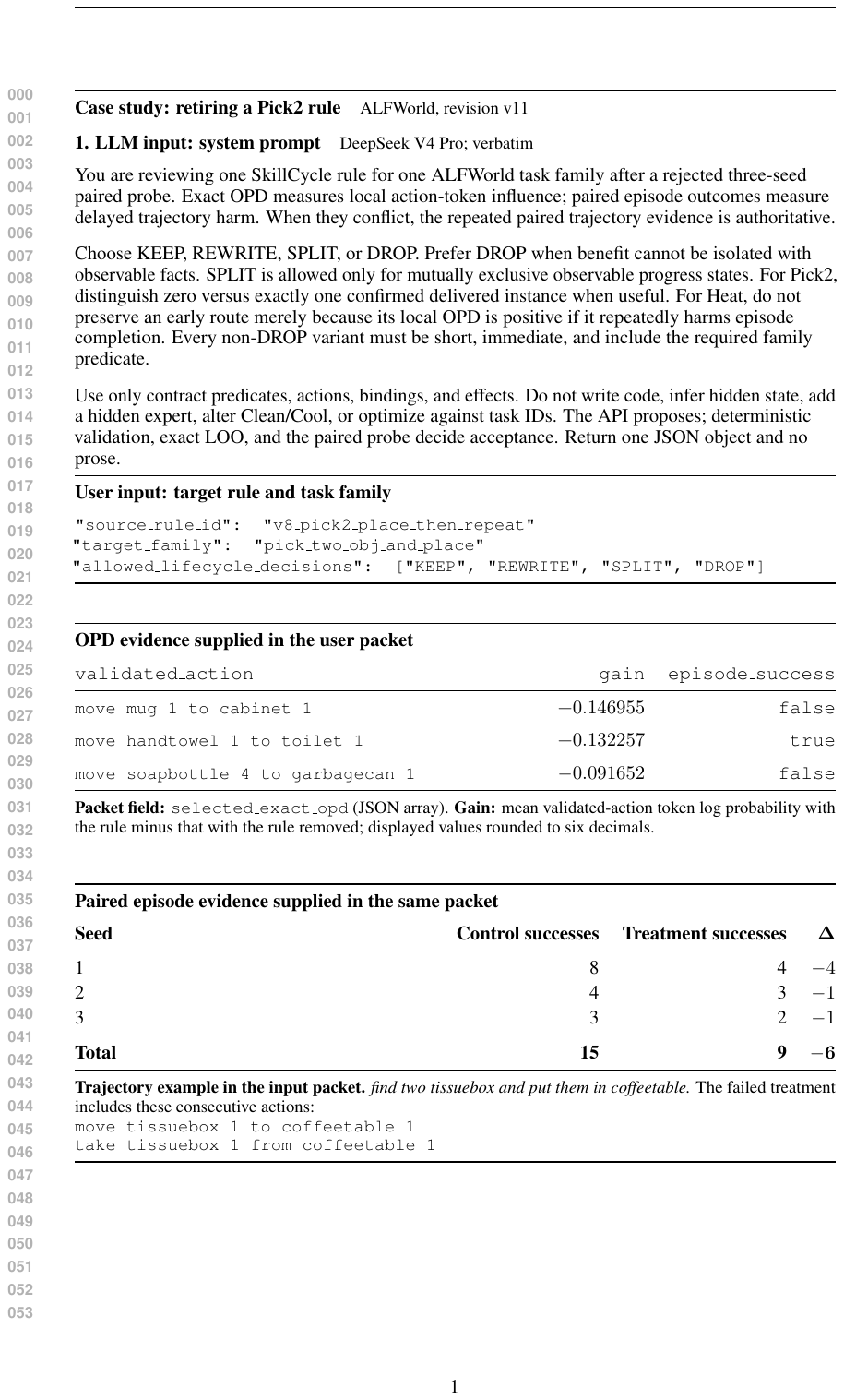}
\par\nointerlineskip
\caption{Pick2 rule retirement: input prompt, OPD evidence, and paired outcomes.}
\label{fig:skill-case-study}
\end{figure}

Figure~\ref{fig:skill-case-study} shows the trajectory feedback and rule
state change for \texttt{v8\_\allowbreak pick2\_\allowbreak place\_\allowbreak then\_\allowbreak repeat}. A failed trajectory
retrieved the same tissuebox after delivery despite the rule's prohibition.
The supplied review packet reports 15 control successes versus nine treatment
successes across three seeds, despite positive local action-token influence
in some examples. The LLM returns \textsc{Drop}; the recorded update
sets the rule inactive while retaining its text in the archive. The prompt's
\textsc{Split} option belongs to this historical review; the current lifecycle
operators are defined in Section~\ref{sec:screening}. Whole-bank acceptance
is assessed separately from this local rule-state change.

\begin{figure}[!htbp]
\centering
\includegraphics[page=2,width=.94\linewidth,trim=106bp 312bp 106bp 80bp,clip]{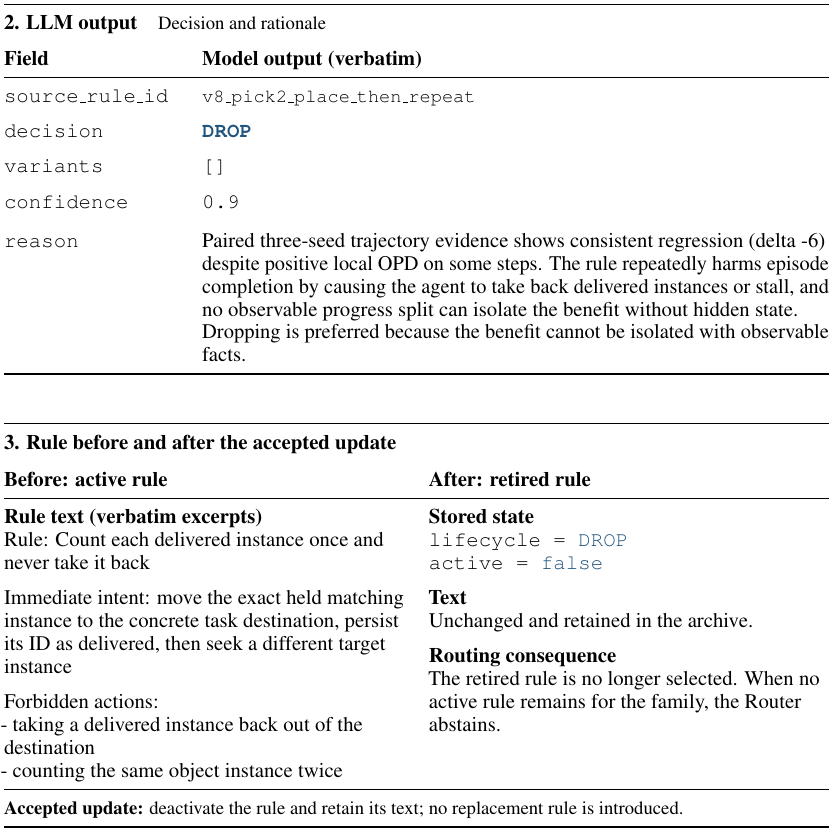}
\par\nointerlineskip
\makeatletter
\@makecaption{Figure~\ref{fig:skill-case-study} (continued)}{LLM decision and rule state before and after retirement.}
\makeatother
\end{figure}
\FloatBarrier

\subsection{Observable Routing and Agent Actions}
In Case A, repeated examination reveals no progress or target object;
only search is eligible, despite its lower priority than take/open rules.
In Case B, cooling is confirmed and the microwave is closed, leaving
opening as the sole eligible rule. An earlier cooling candidate is followed
by an opening action: router candidates provide guidance, not enforced
commands. Table~\ref{tab:historical-router-cases} retains all three seeds.

\begin{table}[!htbp]
\centering\small
\caption{Outcomes and action counts in two routing cases
(F: failure; S: success).}
\label{tab:historical-router-cases}
\begin{tabularx}{\linewidth}{Xccc}
\toprule
Task & Seed & No Skill & Router + Skill\\
\midrule
A: clean soapbar, place in cabinet & 1 & F / 30 & S / 10\\
 & 2 & F / 30 & S / 11\\
 & 3 & F / 30 & S / 11\\
\midrule
B: cool cup, place in microwave & 1 & F / 30 & S / 8\\
 & 2 & F / 30 & S / 8\\
 & 3 & S / 13 & S / 15\\
\bottomrule
\end{tabularx}
\end{table}

\paragraph{Case interpretation.}
Routing resolves the repeated failures in Case A and the first two seeds
of Case B. In Case B's third seed, both conditions succeed, but routing
increases the action count from 13 to 15: useful guidance need not reduce
interaction cost on every trajectory. Eligibility is reconstructed from
the corresponding routing code. These case outcomes are descriptive;
complete checkpoint and prompt identities are unavailable. Controlled revision
comparisons appear in Table~\ref{tab:ablation-core}. The v11 bank failed
whole-bank acceptance; a later combined revision passed its own criteria.
Local rule retirement therefore differs from whole-bank commitment.
Supplementary material contains 226 action--observation pairs and 11 skill documents.
\end{document}